\documentclass[lettersize,journal]{IEEEtran}
\usepackage{array}
\usepackage{multicol}
\usepackage[caption=false,font=normalsize,labelfont=sf,textfont=sf]{subfig}
\usepackage{textcomp}
\usepackage{stfloats}

\usepackage{url}
\usepackage{verbatim}
\usepackage{graphicx}
\usepackage{tikz}
\usepackage{graphicx}
\usepackage{cite}
\usepackage{multirow}
\usepackage{makecell}

\usepackage[ruled,linesnumbered]{algorithm2e}
\usepackage{amsmath}
\usepackage{bm}
\usepackage{color}
\usepackage{graphicx}
\usepackage{amssymb}
\usepackage{booktabs}
\usepackage{multirow}
\usepackage[normalem]{ulem}
\usepackage{ragged2e}
\usepackage{soul}
\usepackage{xcolor}         
\usepackage{colortbl}
\useunder{\uline}{\ul}{}
\usepackage[colorlinks,linkcolor=blue,citecolor=blue]{hyperref}
\usepackage{longtable}
\usepackage{relsize}
\usepackage{comment}
\usepackage{diagbox}

\begin{document}
\title{EvoSampling: A Granular Ball-based Evolutionary Hybrid Sampling with Knowledge Transfer for Imbalanced Learning}

\author{Wenbin Pei*, Ruohao Dai*, Bing Xue,~\IEEEmembership{Fellow,~IEEE,}	Mengjie Zhang,~\IEEEmembership{Fellow,~IEEE}, Qiang Zhang,
Yiu-Ming Cheung,~\IEEEmembership{Fellow,~IEEE}, Shuyin Xia
    \thanks{
    This work was supported in part by the National Key Research and Development Program of China under grant 2021ZD0112400, the National Natural Science Foundation of China under grant 62206041, and the NSFC-Liaoning Province United Foundation under grant U1908214, the 111 Project under grant D23006, the Liaoning Revitalization Talents Program under grant XLYC2008017, and China University Industry-University-Research Innovation Fund under grants 2022IT174, Natural Science Foundation of Liaoning Province under grant 2023-BSBA-030, and an Open Fund of National Engineering Laboratory for Big Data System Computing Technology under grant SZU-BDSC-OF2024-09. (Corresponding author: Wenbin Pei.)
}

   \thanks{Wenbin Pei, Ruohao Dai, and Qiang Zhang are with the School of Computer Science and Technology, Dalian University of Technology, Dalian 116024, China; Key Laboratory of Social Computing and Cognitive Intelligence (Dalian University of Technology), Ministry of Education, Dalian 116024, China (e-mail: peiwenbin@dlut.edu.cn; dairuohao@mail.dlut.edu.cn; zhangq@dlut.edu.cn).

   Bing Xue and Mengjie Zhang are with the School of Engineering and Computer Science, Victoria University of Wellington, PO Box 600, Wellington 6140, New Zealand (e-mails: bing.xue@ecs.vuw.ac.nz; and mengjie.zhang@ecs.vuw.ac.nz)

   Yiu-Ming Cheung is with the Department of Computer Science, Hong Kong Baptist University, Hong Kong, SAR, China (e-mail: ymc@comp.hkbu.edu.hk)

    Shuyin Xia is with the Chongqing Key Laboratory of Computational Intelligence, Key Laboratory of Big Data Intelligent Computing, Key Laboratory of Cyberspace Big Data Intelligent Security, Ministry of Education, Chongqing University of Posts and Telecommunications, Chongqing
400065, China (e-mail: shuyxia@163.com).

*Both authors contributed equally to this research.}
}    

\maketitle
\begin{abstract}
Class imbalance would lead to biased classifiers that favor the majority class and disadvantage the minority class. Unfortunately, from a practical perspective, the minority class is of importance in many real-life applications. Hybrid sampling methods address this by oversampling the minority class to increase the number of its instances, followed by undersampling to remove low-quality instances. However, most existing sampling methods face difficulties in generating diverse high-quality instances and often fail to remove noise or low-quality instances on a larger scale effectively. This paper therefore proposes an evolutionary multi-granularity hybrid sampling method, called EvoSampling. During the oversampling process, genetic programming (GP) is used with multi-task learning to effectively and efficiently generate diverse high-quality instances. During the undersampling process, we develop a granular ball-based undersampling method that removes noise in a multi-granular fashion, thereby enhancing data quality. Experiments on 20 imbalanced datasets demonstrate that EvoSampling effectively enhances the performance of various classification algorithms by providing better datasets than existing sampling methods. Besides, ablation studies further indicate that allowing knowledge transfer accelerates the GP's evolutionary learning process.
\end{abstract}

\begin{IEEEkeywords}
Imbalanced Date Classification, Sampling, Genetic Programming, Multi-task Learning, and Granular Balls Computing.
\end{IEEEkeywords}

\section{Introduction}
\IEEEPARstart{D}{ata} from many practical applications is naturally imbalanced, where the number of instances from different classes varies significantly \cite{he2009learning}. When learning from such data, a seemingly ``intelligent'' classifier deceptively predicts all instances as belonging to the majority class ({\bf{Maj}}) \cite{he2009learning}, thereby achieving high overall accuracy. Unfortunately, such a high overall accuracy is not due to its actual classification ability, but rather the disproportionately larger number of instances in the majority class compared to the minority class ({\bf{Min}}) \cite{10491302}. Moreover, the high overall accuracy makes no sense if {\bf{Min}} is the class of great interest in a classification task \cite{8890005}. For example, earthquakes occur much less frequently than common events, but an earthquake prediction system is specifically designed to detect earthquakes accurately. 




For imbalanced classification, data re-balancing techniques mainly include oversampling, undersampling, and hybrid sampling \cite{he2009learning,pei2023survey}. Undersampling \cite{liu2008exploratory} reduces the number of instances from {\bf{Maj}} to achieve a balanced data distribution across different classes. However, in the case of severe imbalance, significant instances from {\bf{Maj}} are likely to be removed mistakenly \cite{li2019entropy}. Conversely, oversampling methods aim to increase the number of instances in {\bf{Min}} by replicating or generating new instances. Synthetic minority oversampling technique (SMOTE) and its variants \cite{chawla2002smote,han2005borderline,he2008adasyn} are popular in this category. However, these oversampling methods may generate noisy instances, which can lead classifiers to overfit \cite{ren2023grouping,spelmen2018review}.
Differently, hybrid sampling \cite{lu2016hybrid,lin2023towards
} takes advantage of undersampling and oversampling. It typically generates synthetic instances using oversampling techniques (e.g., SMOTE), and subsequently removes low-quality ones by undersampling. However, the following two challenges remain to hybrid sampling.

\begin{figure}
	\centering 
	\includegraphics[ width=0.48\textwidth]{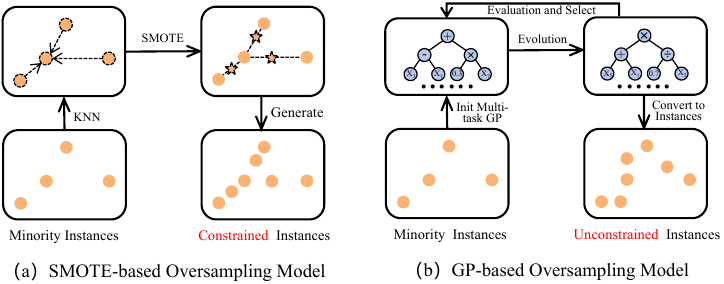}
	\caption{The two types of oversampling models. (a) SMOTE-based approaches require predefining the structure and obtaining neighborhood information to generate an instance between two points. (b) GP-based approaches do not require a predefined structure and can adaptively generate unconstrained instances.} 
	\label{fig:motivation}
\end{figure}

\textbf{Challenge 1:} How to effectively and efficiently generate diverse high-quality instances during the oversampling process? Mainstream hybrid sampling methods \cite{batista2004study} are based on SMOTE to oversample the minority class.
As indicated in Fig.~\ref{fig:motivation} (a), SMOTE generates new instances by a pre-defined linear structure and neighborhood information around a certain {\bf{Min}} instance. However, these generated instances are confined to the space between the target instance and its neighbors, failing to generate diverse instances. Therefore, it is important to develop a method to generate diverse high-quality instances. 

\textbf{Challenge 2:} How to effectively remove low-quality or noisy instances during the undersampling process? 
The predominant undersampling methods~\cite{lin2017clustering,beckmann2015knn} are in reality single-granularity approaches because they do not process data in a multi-granular manner.    

However, in datasets with complex data distributions, single-granularity approaches focus solely on local information, which may result in the loss of key instances or incomplete removal of noise due to their inability to capture the global structure \cite{zhou2012ensemble}. Therefore, it is necessary to develop a multi-granularity undersampling method to address these limitations.

To address \textbf{Challenge 1}, we utilize genetic programming (GP) \cite{poli2008field} to construct high-quality instances. 
In GP, an individual (called a program) is structured by a tree, with nodes selected from predefined terminal and function sets. Each program can be transformed into a mathematical expression, which is treated as an instance and evaluated by a fitness function. 
Moreover, multi-task learning \cite{zhang2021survey} enables GP to evolve diverse instances based on specific target original instances across different tasks and subsequently handle them simultaneously. Across different tasks, their goal of generation tasks may align within the feature space, creating opportunities for knowledge transfer, as indicated in Fig.~\ref{fig:motivation} (b). During the evolutionary learning process of GP, knowledge transfer strategies can capture and transfer reusable knowledge (such as tree structures or key instance characteristics) between different populations corresponding to similar tasks, thereby enhancing both efficiency and effectiveness \cite{9861686}.

To address \textbf{Challenge 2}, we introduce granular ball computing (GBC) \cite{9139397} to remove low-quality or noisy instances. GBC is a data processing method, which processes a complex dataset by dividing the data into multiple granular balls. Unlike traditional sampling methods, GBC constructs granular ball structures that effectively preserve global data characteristics while filtering out noise and outliers through multi-granular adjustments. This method minimizes data loss while retaining essential information. Besides, the construction of granular balls is independent of specific data distributions, making GBC adaptable to various data types and distributions.
These advantages give GBC significant potential for effective undersampling.

The main contributions of this paper are as follows:

\begin{itemize}

\item We introduce a multi-task GP-based oversampling method that incorporates a knowledge transfer strategy to enhance the performance of GP by facilitating knowledge exchange between populations across different tasks. This method operates independently of predefined linear structures and neighborhood information.


\item We propose a GBC-based undersampling method that conducts multi-granularity undersampling at various granular ball levels, effectively removing noisy instances from the dataset. 

\item We propose a hybrid sampling method that combines GP and GBC. Experimental results demonstrate the effectiveness of this hybrid sampling method compared with
existing popular sampling techniques for imbalanced classification.  

\end{itemize}

\section{Background and related works}

\subsection{Genetic Programming}
GP is a heuristic algorithm from the field of evolutionary computation (EC). Inspired by Darwin's theory of evolution, GP evolves programs to solve specific problems through natural selection and genetic operations such as crossover and mutation \cite{poli2008field}. In GP, programs are typically represented using a tree structure, where leaf nodes are selected from a set of terminals, and non-leaf nodes are chosen from a set of functions. The size and structure of programs in GP can dynamically change during evolution, increasing the flexibility of the tree structure. 
\begin{figure}[h]
    \Centering
	\includegraphics[
 width=0.42\textwidth]{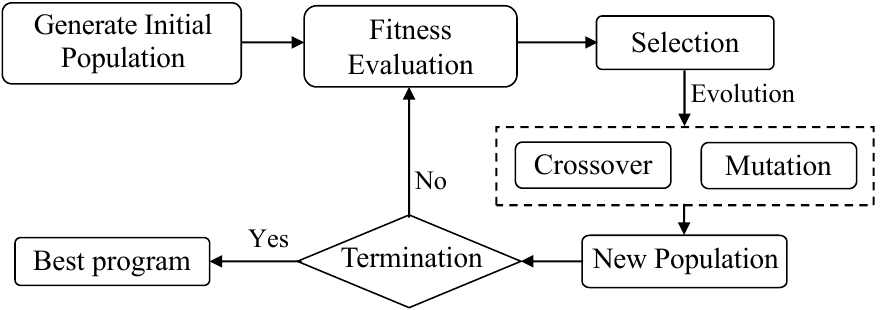}
	\caption{The flowchart of GP.} 
	\label{fig:GP}
\end{figure}
Fig.~\ref{fig:GP} indicates the flowchart of GP. After generating the initial population, the fitness function is used to evaluate how well a program performs. Selection is performed to choose individuals with higher fitness values, and genetic operations are applied to breed offspring in the next generation. GP improves the fitness of individuals through multiple generations of evolution, ultimately identifying a program that effectively solves the problem. 

\subsection{Evolutionary Multi-task Learning}
Multi-task learning (MTL) \cite{zhang2021survey} aims to handle multiple tasks simultaneously to enhance the model's generalization ability. Unlike single-task learning, MTL enhances learning efficiency and model capability by sharing model parameters, feature representations, etc., across related tasks.
Evolutionary multi-task learning (EMTL) \cite{ong2016evolutionary} combines EC with MTL to address multiple tasks simultaneously in their search space. Different tasks promote each other by sharing relevant knowledge, thereby improving search efficiency and learning performance. In general, EMTL performs particularly well in situations with limited data or high complexity \cite{9868257, 8944273}. The similarity between tasks is important to evaluate the effectiveness of multi-task learning. When tasks are highly correlated, the impact of knowledge transfer becomes more significant. 
By sharing information between tasks, simpler tasks could accelerate the learning of more complex tasks, while the features of complex tasks could, in turn, enhance the performance of simpler tasks.

A representative in EMTL is the multi-factorial evolutionary algorithm (MFEA) \cite{gupta2015multifactorial}. In MFEA, individuals are assigned to tasks based on a skill factor $\tau$, allowing gene sharing and cross-task crossover to leverage knowledge transfer. 
The subsequent MFEA-II \cite{bali2019multifactorial} refines this mechanism further by dynamically controlling the degree of knowledge transfer according to task similarity. This ensures that related tasks could more effectively benefit from each other. Building on these foundational methods, other EMTL frameworks have also demonstrated the effectiveness of multi-task optimization. For instance, evolutionary multi-task incremental learning (EMTIL) \cite{8944273} enhances knowledge transfer by selecting transfer strategies using incremental classifiers, ensuring that the most useful information is shared between tasks. In addition, multi-population approaches, such as multi-objective multifactorial GP (MOMFGP) \cite{zhang2022multitask} and adaptive transfer multi-task GP (ATMTGP) \cite{9868257}, further improve EMTL by facilitating inter-task crossovers, where knowledge is exchanged between tasks, with the degree of transfer dynamically adjusted based on task similarity.

\subsection{Granular Ball Computing}

\begin{figure}
    \Centering
	\includegraphics[width=0.4\textwidth]{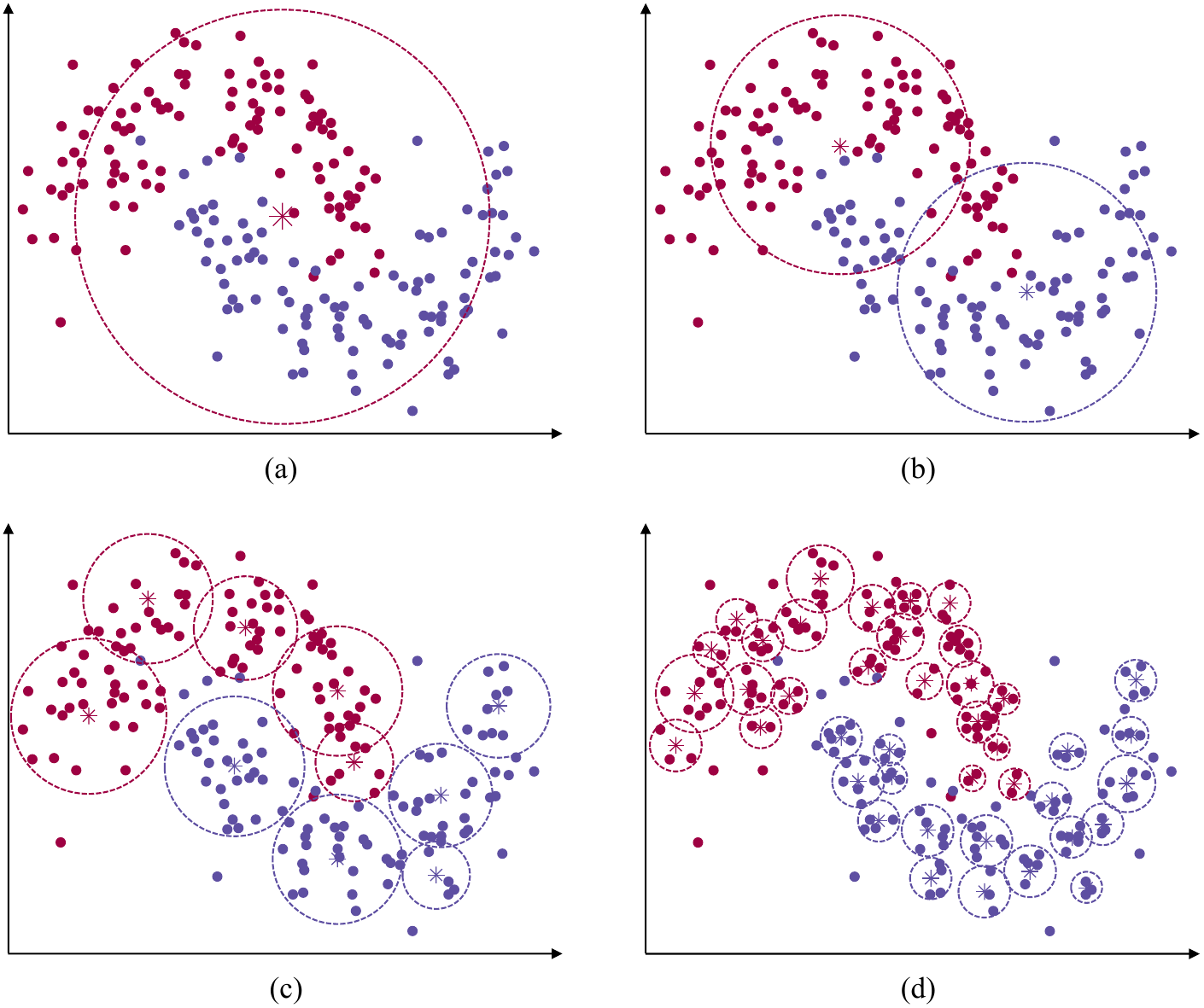}
	\caption{The generation process of GBs. (a) The initial GB. (b) and (c) GBs in the intermediate process. (d) The final GBs.} 
	\label{fig:GB}
\end{figure}
GBC \cite{9139397} is a promising
data processing and analysis method based on granular computing. It has been successfully applied across various domains, including granular ball classifiers \cite{xia2022efficient}, and granular-ball clustering \cite{10530064}. As shown in Fig.~\ref{fig:GB}, the core idea of GBC is to partition data points into granular balls (GB) of varying granularity, where balls of different sizes cover data to enhance its representation, processing, and analysis. A granular ball covers $n$ instances, i.e., GB = $\{x_1, ..., x_n\}$. The center $c$ and radius $r$ of a GB are defined as follows \cite{xia2022efficient}:

\begin{equation}
\label{GB_center}
c = \frac{1}{N}\sum_{i = 0}^N x_{i},
\end{equation}

\begin{equation}
\label{GB_radius}
r = \frac{1}{N}\sum_{i = 0}^N |x_{i}-c|,
\end{equation}

where $x_i$ is an instance, and $N$ is the number of instances covered by a GB. 


The quality of GB, denoted as $T_{GB}$, is typically calculated based on the purity or distribution of the instances within the ball. For using GB to imbalanced classification, the quality of a GB can be measured by:


\begin{equation}
\label{GB_quality}
T_{GB} = \frac{\max(|N_i|)}{\sum |N_i|},   i = 1, 2,...,k,
\end{equation}
where $|N_i|$ is the number of instances in class $i$ inside the GB. The closer the quality value to 1, 
the more homogeneous the instances within the granular ball. Given a quality threshold $T$, if $T_{GB}$ $< T$, the ball must be split until all GBs satisfy the threshold. 
Through this process, the instance categories within GB become mostly the same, allowing the GB to represent the category distribution of the instance data more accurately.

\subsection{Existing Sampling Methods in Imbalanced Classification}

Random undersampling (RUS) randomly removes a portion of {\bf{Maj}} to balance data, which often leads to the loss of important instances. To more effectively retain valuable {\bf{Maj}}, the NearMiss method \cite{mani2003knn} introduces a strategy that selects {\bf{Maj}} instances closer to {\bf{Min}} and removes those that are far away. Although NearMiss theoretically preserves useful information, it may still cause the loss of essential majority class data, ultimately impacting the classifier’s performance. Some studies propose using clustering techniques \cite{lin2017clustering,ofek2017fast} and ensemble learning \cite{REN2022108295,liu2009exploratory} with undersampling.

For oversampling, SMOTE is widely used. SMOTE generates new instances by selecting a {\bf{Min}} instance $x_i$ and one of its k-nearest neighbors $x_j$, using the following formula:

\begin{equation}
\label{smote_eq}
x = x_{i} + (x_{j} - x_{i}) \times rand(0,1),
\end{equation}
where $rand(0,1)$ is a random number distributed between 0 and 1.

However, a major limitation of SMOTE and its improved variants \cite{he2008adasyn,zhu2017synthetic,chen2010ramoboost,ren2017ensemble} is their reliance on linear interpolation for synthesizing new instances. This restricts the diversity and variability of generated instances \cite{krawczyk2016learning}. Furthermore, linear interpolation can blur the boundary between {\bf{Maj}} and {\bf{Min}}, particularly in overlapping regions, which often results in noisy instances and unclear class boundaries \cite{ren2023grouping}.

Hybrid sampling methods such as SMOTE+ENN \cite{batista2004study} and SMOTE+Tomek \cite{batista2004study} are introduced to address these issues. In these methods, SMOTE is first used to generate synthetic instances, followed by undersampling techniques, such as edited nearest neighbor (ENN) or Tomek Links, to remove noisy data. These hybrid methods improve the data quality by removing noise introduced by SMOTE. Nevertheless, these methods with linear interpolation make it hard to generate diverse instances. 
Moreover, these methods are all single-granularity undersampling, which may perform less effectively when dealing with data with intricate class boundaries. They may remove important instances in the boundary, resulting in information loss.


\section{The Proposed Method}
\subsection{The Framework of EvoSampling}
\begin{figure*}
	\centering 
	\includegraphics[width=1\textwidth]{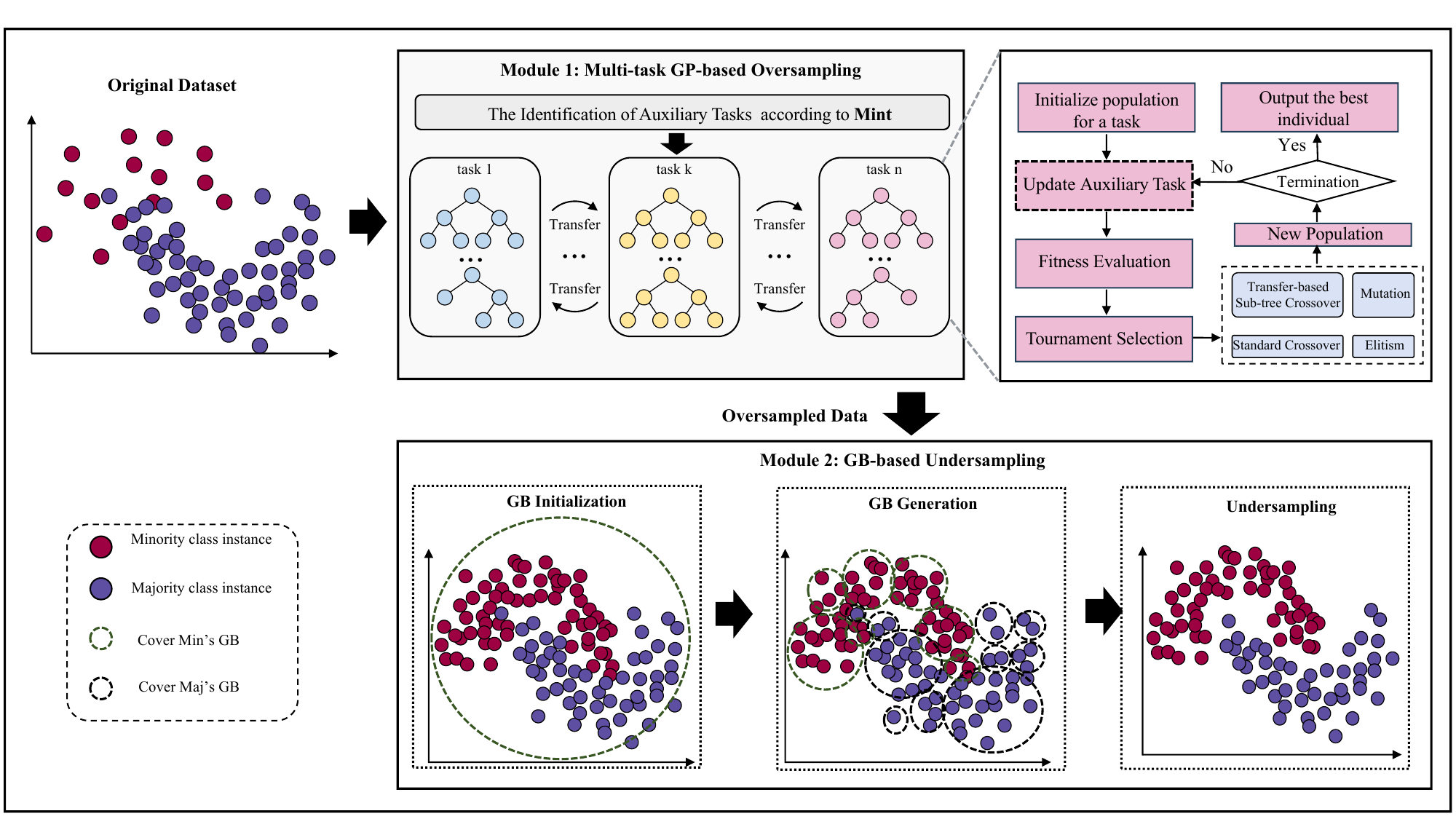}
	\caption{The framework of EvoSampling. The original dataset is oversampled using the multi-task GP to balance data. The GBC is then used to perform multi-granularity undersampling on the oversampled data to remove low-quality instances.} 
	\label{fig:framework} 
\end{figure*}
The EvoSampling method, as illustrated in Fig.~\ref{fig:framework}, consists of two components: (1) a GP-based oversampling model with knowledge transfer operations, and (2) a GBC-based undersampling model.

In the GP-based oversampling learning process, to generate $n$ instances for the minority class, $n$ GP processes are required, each of which is executed in parallel for a unique task aiming to generate an instance based on a target {\bf{Maj}} and {\bf{Min}} instances. EvoSampling assigns an auxiliary task to each GP process based on the similarity between tasks. After initializing the population of a GP process, individuals within the population are evaluated by the fitness function. Note that during evaluations, different GP processes use distinct target {\bf{Maj}} and {\bf{Min}} instances based on their assigned tasks. 
The tournament selection selects superior individuals based on fitness values. In addition to using the standard sub-tree crossover and mutation to breed offspring, a transfer-based crossover operator is designed, allowing the selected individuals to cross not only with others within their population but also with elite individuals from the auxiliary task. This facilitates the transfer of high-quality knowledge, speeding up the convergence of the evolutionary process. The $n$ GP processes are conducted in parallel with knowledge transfer, each outputting its best individual at the end of the learning process.


The second component comprises three major steps, i.e., GB initialization, GB generation, and GB-based undersampling. During the GB initialization process, a granular ball contains all the instances after oversampling. During the GB generation process, low-quality balls are split until each reaches the desired quality level, creating multiple high-quality balls of varying granularity covering all the instances. During the undersampling process, the centers of the generated granular balls are the basis for determining whether to remove a ball, depending on the class distribution of neighboring instances. 

 

\subsection{The Multi-task GP-based Oversampling Method}

\subsection*{1) Representation of a GP individual}
Due to the flexible tree structure, GP is capable of producing diverse and higher-quality instances. In EvoSampling, a GP individual works as a synthetic instance. To achieve this, the terminal set is designed to include all the instances from the minority class. To operate on terminal nodes, the function set includes four arithmetic operators, i.e., $+$, $-$, $\times$, and $\div$ (protected). The protected $\div$ function returns 1 when the denominator is 0.

\begin{figure}
	\centering 
	\includegraphics[ width=0.25\textwidth]{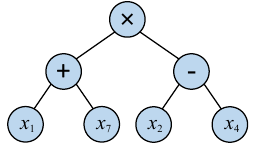}
	\caption{A synthetic instance represented by a GP individual.} 
	\label{fig:gp_example}
\end{figure}

By converting a GP individual into a mathematical expression, a new instance is synthesized automatically. For example, the GP individual shown in Fig.~\ref{fig:gp_example} converts to a new instance: $x_{new} = (x_{1} + x_{7}) \times (x_{2} - x_{4})$, where $x_{1}$, $x_{7}$, $x_{2}$, and $x_{4}$ are good-quality instances selected from the terminal set (i.e., the minority class). Therefore, the GP-based instance generation is independent of any pre-defined formulation.

Moreover, GP has a built-in instance selection capability. This is because the fitness function evaluates individuals in a population to drive the evolutionary process toward better solutions. As a result, individuals with poor-quality instances as leaf nodes 
are more likely to have low fitness values, reducing their chances of survival.

\subsection*{2) The Identification of Auxiliary Tasks}

In EvoSampling, $n$ GP processes are required to generate $n$ instances for the minority class. Each process addresses a task of generating an instance based on a specific pair of target {\bf{Maj}} and {\bf{Min}} instances, denoted as ({\bf{Maj$_t$}}, {\bf{Min$_t$}}). Note that various tasks use distinct ({\bf{Maj$_t$}}, {\bf{Min$_t$}}) to ensure that the generated $n$ instances are diverse. 


Before evolutionary learning, EvoSampling allocates ({\bf{Maj$_t$}}, {\bf{Min$_t$}}) to each task, based on the following steps:
\begin{itemize}	
	\item \textbf{Step 1:} The centers of {\bf{Maj}} and {\bf{Min}} are identified by calculating the mean value on every dimension, denoted as {\bf{Maj$_c$}} and {\bf{Min$_c$}}, respectively.	
	\item \textbf{Step 2:} Euclidean distance from each {\bf{Maj}} instance to {\bf{Maj$_c$}} is calculated to identify the nearest $n$ as {\bf{Maj$_t$}} instances. 
	\item \textbf{Step 3:} If {\bf{$|$Min$|$}}$\geq n$, the nearest $n$ instances to {\bf{Min$_c$}} are identified as the $n$ target {\bf{Min}} instances, each of which is sequentially used with a {\bf{Maj$_t$}} instance to form ({\bf{Maj$_t$}}, {\bf{Min$_t$}}). If {\bf{$|$Min$|$}}$< n$, all the  {\bf{Min}} instances must be used as {\bf{Min$_t$}} more than once, i.e., multiple {\bf{Maj$_t$}} share the same {\bf{Min$_t$}}. 
\end{itemize}


In the proposed multi-task GP-based oversampling method, multiple tasks are performed simultaneously, allowing useful knowledge to transfer between them to accelerate the learning process. However, when tasks may differ significantly, knowledge transfer may result in performance loss due to negative transfer. Therefore, it is crucial to identify an auxiliary task that is highly relevant to the target task.

Before the evolutionary learning process, task relatedness can be measured by comparing differences in their ({\bf{Maj$_t$}}, {\bf{Min$_t$}}). All tasks have distinct {\bf{Maj$_t$}},
while some share the same {\bf{Min$_t$}} when the class imbalance ratio $IR>2$ (where $IR$ is equal to the number of {\bf{Maj}} instances divided by the number of {\bf{Min}} instances). Therefore, tasks with the same {\bf{Min$_t$}} become a group, denoted $G= \{T_1, T_2,\dots,T_h\}$. Afterward, for each task within a group, its auxiliary task is identified by comparing the distance between its {\bf{Maj$_t$}} and those of the other tasks within that group. Specifically, for $T_i \in G$, we calculate the Euclidean distance between its {\bf{Maj$_t^i$}} and each {\bf{Maj$_t^j$}} (i.e., the {\bf{Maj$_t$}} of task $T_j$ within the group $G$). After calculating distances between {\bf{Maj$_t^i$}} and {\bf{Maj$_t$}} values of all the other tasks in group $G$, the task with the shortest distance is identified as the auxiliary task for $T_i$.

However, a task's auxiliary task may dynamically change as new useful knowledge is acquired throughout the evolutionary learning process. The best individual from a population is a representative solution to addressing the targeted task. Hence, EvoSampling calculates the distances between the best individuals from all the $n$ GP processes associated with $n$ tasks to dynamically determine auxiliary tasks during the evolutionary process. For example, for task $T_i \in G$ addressed in the $ith$ GP process, EvoSampling computes the Euclidean distance of its best individual to each of the best individuals evolved for the other tasks within $G$, to identify the closest one as its auxiliary task. The auxiliary task to the targeted task is updated every 10 generations to save computational costs. 

\subsection*{3) The Fitness Function}

\begin{figure}
	\centering 
	\includegraphics[ width=0.35\textwidth]{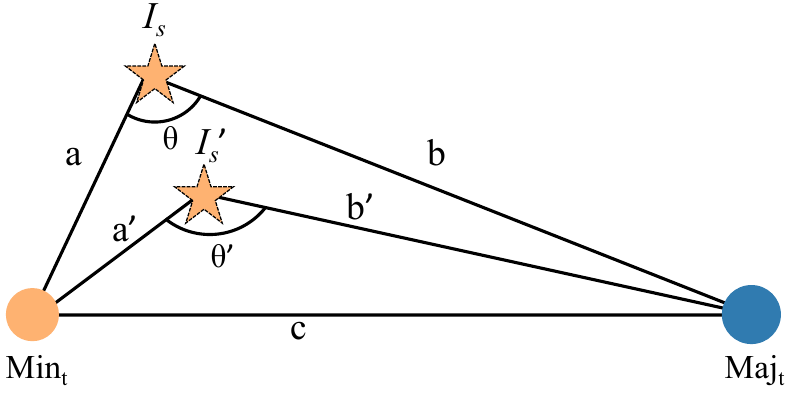}
	\caption{Different angles of two different generated instances $I_s$ and $I_s'$ with {\bf{Maj$_t$}} and {\bf{Min$_t$}}.}
 \label{fig:instance}
\end{figure}

In EvoSampling, an individual is treated as a generated $I_s$. 
The three points in the space, i.e., $I_s$, {\bf{Maj$_t$}} and {\bf{Min$_t$}}, form a triangle, as indicated in Fig. \ref{fig:instance}. In the triangle, $a$ represents the distance between $I_s$ and {\bf{Min$_t$}}, $b$ represents the distance between $I_s$ and {\bf{Maj$_t$}}, and $c$ is the distance between {\bf{Min$_t$}} and {\bf{Maj$_t$}}. A generated synthetic instance $I_s$ is considered high-quality if its distance to {\bf{Min$_t$}} is shorter than to that of {\bf{Maj$_t$}}, i.e., $a<b$. This can be mathematically measured by:

\begin{equation}
\label{distance}
D = \frac{e-e^{\frac{a}{a + b}}}{e - 1},
\end{equation}
where $e$ is the base of the natural logarithm. The design details of distance $D$ are introduced in the \textbf{supplementary materials}. Note that $a<b$ is equivalent to $D >\frac{e-e^{0.5}}{e-1}$, explained in \textbf{supplementary materials}. 

In the triangle, according to the cosine theorem, the angle $\theta$ between $a$ and $b$ is calculated:

\begin{equation}
\label{distance}
\theta = arccos[\frac{a^2 + b^2 -c^2}{2ab})].
\end{equation}
The angle $\theta$ ranges from 0 to $180^\circ$.
Note that the angle measure is meaningful if $D >\frac{e-e^{0.5}}{e-1}$. In that case, a larger $\theta$ is preferred. As indicated in Fig. \ref{fig:instance}, a larger $\theta'$
indicates that a generated instance is closer to {\bf{Min$_t$}} than to {\bf{Maj$_t$}}. Using angle $\theta$ further benefits capturing directional relationships and encourages generated instance $I_s$ to move towards {\bf{Min$_t$}}. Therefore, to comprehensively evaluate the quality of the generated instances, both the distance and angle are considered to design the fitness function as follows:

\begin{equation}
\label{fitness}
Fitness = [D, \theta].
\end{equation}

\subsection*{4) Selection}
Based on obtained fitness values, better individuals are selected by the tournament selection with the tournament size of $ts$. Inspired by \cite{pei2022high}, after randomly selecting $ts$ individuals from the population (denoted as $Ind_{ts}$), the selection process in a tournament is as follows:
\begin{itemize}

     \item \textbf{Step 1:} 
     From the $Ind_{ts}$, we select individuals with $D >\frac{e-e^{0.5}}{e-1}$ (this is equivalent to $a<b$), denoted as $Ind_d$. This process is to exclude low-quality individuals that are closer to {\bf{Maj$_t$}} than to {\bf{Min$_t$}}.

     \item \textbf{Step 2:} From the set $Ind_d$, individuals with the highest $\theta$ value are selected, and then the individual with the highest $D$ value among them is chosen as the winner in the tournament. 

 \end{itemize}

\subsection*{5) Crossover and Mutation}

\begin{figure*}
	\centering 
	\includegraphics[width=1.0\textwidth]{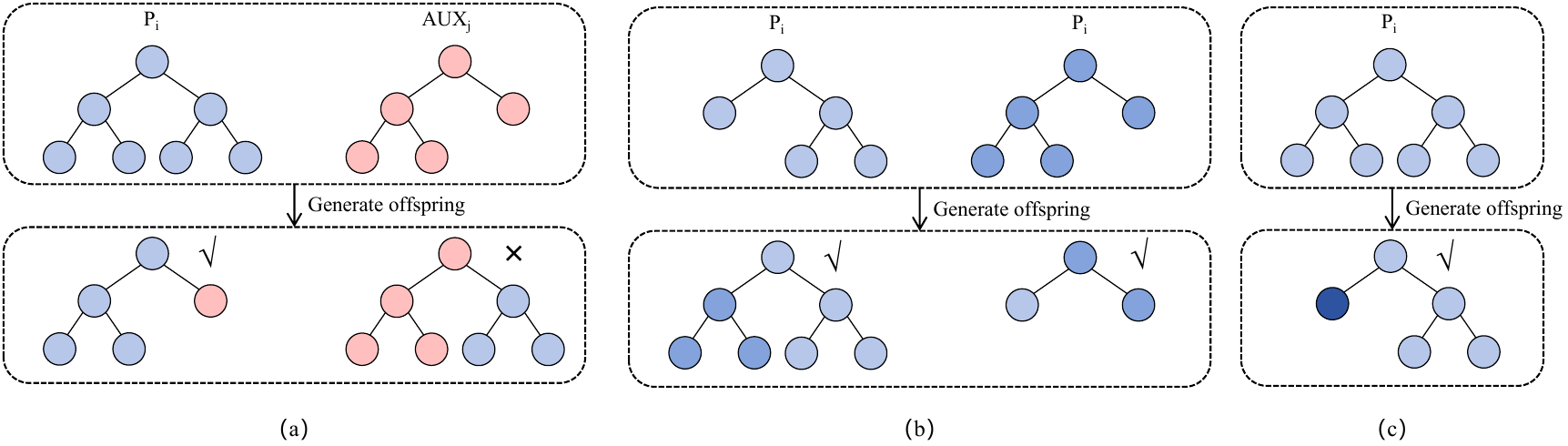}
	\caption{Crossover and Mutation. (a) The transfer-based sub-tree crossover; (b) Standard sub-tree crossover; (c) Standard mutation.
 } 
	\label{fig:c_m}
\end{figure*}

In EvoSampling, two types of crossover operators and a mutation operator are involved. 



\textbf{Transfer-based Sub-tree Crossover}: This operation facilitates knowledge transfer by enabling individuals from the target population (i.e., for the current task to be addressed) to crossover with high-performing individuals from the auxiliary population (i.e., for the auxiliary task), thereby producing potentially superior offspring.

Fig.~\ref{fig:c_m} \textbf{(a)} indicates the transfer-based crossover operator. First, one parent $P_i$ is selected from the target population, while the other parent {AUX}$_j$ is randomly selected from a set of high-performing individuals (i.e., the top 30\% individuals) from the auxiliary population. On each parent, a crossover point is randomly selected, and then the sub-trees rooted at these crossover points are swapped. This produces two new offspring. Among them, one with the root node belonging to the target population is retained, while the other with the root node belonging to the auxiliary population is discarded. This ensures that the target population gains benefits from useful knowledge of the auxiliary population without being substantially changed.

\textbf{Standard Sub-tree Crossover}: 
Fig.~\ref{fig:c_m} \textbf{(b)} indicates the standard crossover operator. 
In the standard crossover, both parents are all from the target population, with two crossover points randomly selected. The subtrees rooted at these crossover points are swapped, generating two new offspring. This operation allows the offspring to inherit genetic information from their parents.

\textbf{Mutation Operator}: 
Fig.~\ref{fig:c_m} \textbf{(c)} indicates the mutation operator. In this mutation operation, an individual from the target population is selected as a parent, with a mutation point randomly selected on it. A new randomly generated sub-tree then replaces the original sub-tree rooted at the mutation point.  


\subsection{The Proposed GB-based Undersampling Method}
Traditional undersampling methods typically operate at a single granular level of data points. However, noise within a dataset often displays diversity in distribution and density, which makes it difficult for single-granularity methods to eliminate it effectively. Introducing GB for undersampling enables the handling of data at multiple granular levels. 
Therefore, in EvoSampling, GBs with varying granularities are generated on the oversampled data, and undersampling is then performed at the GB level to eliminate noise. 
\subsection*{1) The GB Generation}


For the oversampled dataset $\mathcal{D}_1$, a GB set $\mathcal{A}$ = \{$GB_{1}$,...,$GB_m$\} is created, where $m$ represents the total number of generated granule balls. In $\mathcal{A}$, each granule ball potentially contains a different number of instances, reflecting data distributions across various granular levels. The different GB sizes enable data to be undersampled at multiple granular levels.

The process of generating granule balls is described in Fig. \ref{fig:GB}. Initially, as indicated in Fig. \ref{fig:GB} (a), an initial granule ball is created encompassing all the instances in $\mathcal{D}_1$, with its center defined as the center of all the instances. Although the initial GB retains complete dataset information, its granularity is too coarse with a low quality measured by equation ({\ref{GB_quality}}). Consequently, the initial GB is iteratively split until all the split granule balls reach a predefined quality threshold $\mathcal{T}$. 

In this splitting process, for each granule ball $GB_{i}$ in $\mathcal{A}$, its quality $T_{i}$ is calculated according to equation ({\ref{GB_quality}}). If $T_{i}$ $<$ $\mathcal{T}$, the granule ball $GB_{i}$ is split, otherwise it is unchanged. To split $GB_{i}$, a data point $x$ with a class label different from that of $GB_{i}$ is randomly selected from $GB_{i}$. Then, the center of $GB_{i}$ and the data point $x$ are treated as two new centers to create two new GBs, denoted as $GB_{i}^1$ and $GB_{i}^2$, respectively. We then calculate the distance from each point in $GB_{i}$ to the centers of $GB_{i}$ and $x$. Data points closer to $x$ are assigned to $GB_{i}^2$, while the rest are assigned to $GB_{i}^1$. Subsequently, $GB_{i}$ is replaced by $GB_{i}^1$ and $GB_{i}^2$, updating the set $\mathcal{A}$. This process repeats until all $T_{i}$ $\geq$ $\mathcal{T}$.

\subsection*{2) Undersampling based on the Generated GBs}

After generating the GB set, $GB_i$ ($GB_i \in$ $\mathcal{A}$) is represented by the center $c_i$ and radius $r_i$ of the ball. The center $c_i$ represents the position of $GB_i$, and the radius $r_i$ represents the granularity of $GB_i$. To identify the nearest neighbors of $GB_i$, the distance is calculated as follows:

\begin{equation}
\label{distance_ball}
D_{ball} = d(c_i,c_j) - r_i -r_j,
\end{equation}
where $c_i$ is the center of $GB_i$; $c_j$ is the center of $GB_j$; $d(c_i,c_j)$ represents the Euclidean distance between $c_i$ and $c_j$; $r_i$ is the radius of $GB_i$ and $r_j$ is the radius of $GB_j$.

If the labels of all the nearest neighbors of $GB_i$ are the same as that of $GB_i$, then $GB_i$ is unchanged. Conversely, if the label of a ball among the nearest neighbors of $GB_i$ differs from that of $GB_i$, then $s$ instances in $GB_i$ will be randomly removed. Inspired by
SMOTE+Tomek \cite{batista2004study}, the number $s$ is designed as follows:


\begin{equation}
s = \frac{1}{k} \sum_{j=1}^{k}(1-\mathbf{I}\{L_i = L_j\}) \times s_j,
\label{eq_n}
\end{equation}
where $k$ is the number of nearest neighbors, $L_i$ is the label of $GB_i$, $L_j$ is the label of the $j$th ball among its the nearest neighbors, and $s_j$ is the number of instances covered by the $j$th ball. The indicator function $\mathbf{I}\{a = b\}$ equals 1 if $a = b$; otherwise, it is equal to 0.

\begin{figure}
	\centering 
	\includegraphics[width=0.4\textwidth]{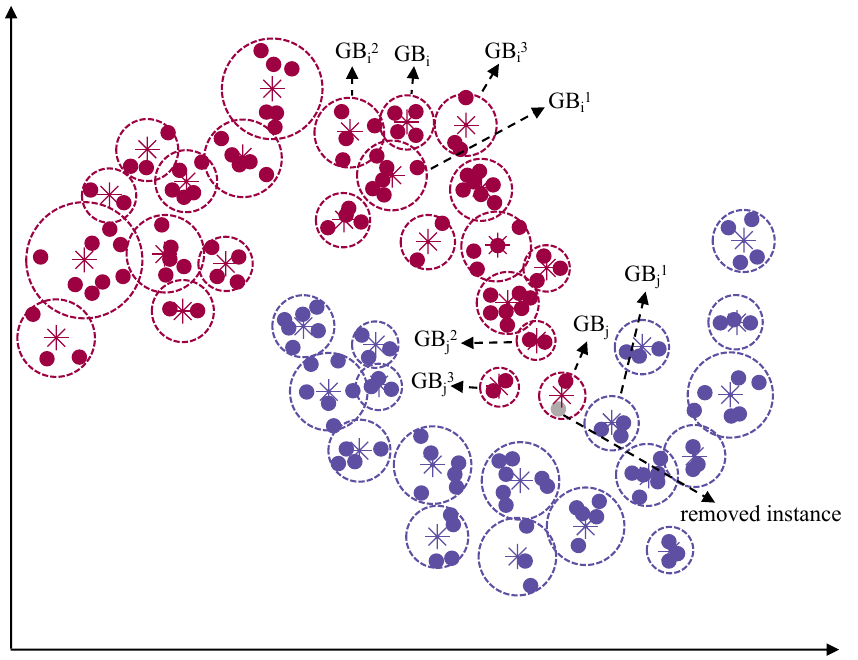}
	\caption{An example of undersampling based on GBs.} 
	\label{fig:undersampling}
\end{figure}

As the example shown in Fig.~\ref{fig:undersampling}, the nearest 3 neighbors of $GB_i$ are $GB_i^1$, $GB_i^2$, and $GB_i^3$. In this example, the 3 neighbors contain instances with the same class label as $GB_i$, then $GB_i$ remains unchanged. However, for $GB_j$, its nearest 3 neighbors are $GB_j^1$, $GB_j^2$, and $GB_j^3$, where the label of $GB_j^3$ is different from $GB_j$. Therefore, $s$ instances (where $s$ is calculated by equation (\ref{eq_n})) are randomly removed from $GB_j$. 

After every granular ball in $\mathcal{A}$ finishes removing instances, the undersampled data is likely to become imbalanced again. In this case, EvoSampling compares the number of instances in each class to identify which class includes a large number of instances. Then, it removes the smallest ball(s) covering the instances from the class. 
This is because small balls are usually located at the boundary or between different classes. The above step repeats until the numbers of the two classes are rebalanced. The pseudocode for the GB-based undersampling is shown in Algorithm \ref{algorithm_2}. 

\begin{algorithm}
    \caption{GB-based Undersanmpling}
    \label{algorithm_2}
    \KwIn{The GB set $\mathcal{A}$;}
    \KwOut{The final undersampled data $\mathcal{D}_2$;}
    \For{$GB_{i}$ in $\mathcal{A}$}
    {
        Calculate the center $c_i$, radius $r_i$, and the label $L_i$\;
    }
    \For{$GB_{i}$ in $\mathcal{A}$}
    {
         Identify nearest neighbors of $GB_i$\; 
         Calculate the labels of nearest neighbors; 

          \uIf{\text{all their labels are same as} $L_i$}
        {
           $GB_{i}$ remains the same;
        }
        \Else
        {
            Randomly remove $s$ instances from $GB_{i}$\;
        }
    }
 
    Calculate $|${\bf{Maj}}$|$ and $|${\bf{Min}}$|$ of undersampled data;
    \While{$|${\bf{Min}}$|$ $ \neq $ $|${\bf{Maj}}$|$}
    {   
        \uIf{$|${\bf{Maj}}$|$ $>$ $|${\bf{Min}}$|$}
        {
            Remove the smallest ball labeled {\bf{Maj}} in $\mathcal{A}$\;
            Update $|${\bf{Maj}}$|$\;
        }
        \Else
        {
            Remove the smallest ball labeled {\bf{Min}} in $\mathcal{A}$\;
            Update $|${\bf{Min}}$|$\;
        }

    }
    Integrate instances $\mathcal{D}_2$ contained by all the GBs in $\mathcal{A}$\;
    return $\mathcal{D}_2$;
\end{algorithm}

\section{Experimental Design}
\subsection{Datasets}

In our experiments,
20 datasets\footnote{These datasets are available at: \href{https://sci2s.ugr.es/keel/imbalanced.php}{https://sci2s.ugr.es/keel/imbalanced.php}} are used to thoroughly evaluate the effectiveness of the proposed method. The details of these datasets are reported in TABLE ~\ref{tab:datasets}.
These datasets are from various domains, including medical diagnosis, industrial production, etc. 
The imbalance ratio $IR$ values of these datasets range from 9.09 to 85.88, covering a wide range of imbalance levels.  Hence, these experiments on these imbalanced datasets can validate the effectiveness of our method in rebalancing imbalanced data, particularly those with high imbalance ratios ($IR \geq 10$). To split data, we employ stratified sampling, allocating 70\% of the data for training and 30\% for testing, ensuring that both sets retain the same $IR$ as the original dataset.

\begin{table}[h]
\caption{Datasets in the experiments}
\label{tab:datasets}
\centering
\begin{tabular}{|c|c|c|c|}
\hline
Dataset & NO. instances & IR    & Index \\ \hline
ecoli-0-6-7\_vs\_3-5         & 222      & 9.09  & D1    \\ \hline
ecoli-0-2-6-7\_vs\_3-5       & 224      & 9.18  & D2    \\ \hline
ecoli-0-3-4-6\_vs\_5         & 205      & 9.25  & D3    \\ \hline
ecoli-0-3-4-7\_vs\_5-6       & 257      & 9.28  & D4    \\ \hline
glass-0-6\_vs\_5             & 108      & 11    & D5    \\ \hline
glass-0-1-4-6\_vs\_2          & 205      & 11.06 & D6    \\ \hline
glass2                       & 214      & 11.59 & D7    \\ \hline
ecoli-0-1-4-7\_vs\_5-6       & 332      & 12.28 & D8   \\ \hline
ecoli-0-1-4-6\_vs\_5         & 280      & 13    & D9   \\ \hline
yeast-1\_vs\_7               & 459      & 14.3  & D10   \\ \hline
glass4                       & 214      & 15.47 & D11   \\ \hline
yeast-1-4-5-8\_vs\_7          & 693      & 22.1  & D12   \\ \hline
glass5                       & 214      & 22.78 & D13   \\ \hline
poker-9\_vs\_7               & 244      & 29.5  & D14   \\ \hline
winequality-white-3\_vs\_7   & 900      & 44    & D15   \\ \hline
winequality-white-3-9\_vs\_5 & 1482     & 58.28 & D16   \\ \hline
poker-8-9\_vs\_6             & 1485     & 58.4  & D17   \\ \hline
winequality-red-3\_vs\_5     & 691      & 68.1  & D18   \\ \hline
poker-8-9\_vs\_5             & 2075     & 82    & D19   \\ \hline
poker-8\_vs\_6                & 1477     & 85.88 & D20   \\ \hline
\end{tabular}
\end{table}

\subsection{Baseline Methods}
We compare our method with 4 popular oversampling methods and 2 popular hybrid sampling methods. These methods are described as follows:
\begin{itemize}
    \item \textbf{SMOTE \cite{chawla2002smote}:} it is the most commonly-used oversampling method that synthesizes new instances by interpolating between a {\bf{Min}} instance and its neighbors to balance data.
    \item \textbf{ADASYN \cite{he2008adasyn}:} He et al. propose an adaptive oversampling method that prioritizes the difficult-to-learn instances in {\bf{Min}} to generate more synthetic instances to balance data.   
    \item \textbf{Borderline-SMOTE1 \cite{han2005borderline}:} It synthesizes new instances by interpolating borderline instances in {\bf{Min}} with their same-class (minority) neighbors.
    \item \textbf{Borderline-SMOTE2 \cite{han2005borderline}:} It synthesizes new instances based on borderline instances in {\bf{Min}}   (labeled as $danger$ instances) and their same-class neighbors. It also considers {\bf{Maj}} neighbors in some cases, creating synthetic instances that tend to be closer to {\bf{Min}} but in a way that represents the challenging decision boundary region more accurately.
    \item \textbf{SMOTE+ENN \cite{batista2004study}:} It combines SMOTE with ENN technique, removing potentially misclassified instances after generating new instances for {\bf{Min}} to balance data.
    \item \textbf{SMOTE+Tomek \cite{batista2004study}:} It combines SMOTE with the Tomek Links cleaning method. SMOTE first generates synthetic instances for  {\bf{Min}}, and then the Tomek Links method removes pairs of instances from different classes that are closest to each other (the ``Tomek Links'').

\end{itemize}

\subsection{Parameter Settings}

\begin{table}[]
\caption{PARAMETER SETTINGS}
\label{tab:parameter}
\centering
\begin{tabular}{c c}
\hline
Parameter                          & Value                \\ \hline
Population size           & 30*($|{\bf{Maj}}|-|{\bf{Min}}|$)        \\  
The number of generations          & 50 \\ 
Tournament size                    & 3                  \\  
Rate of standard crossover  & 50\%                \\ 
Rate of transfer-based crossover            & 30\% \cite{gupta2015multifactorial}                 \\
Rate of mutation                           & 20\%                 \\ 
initialization & Ramped-half-and-half \\
Maximal tree depth        & 10                   \\
Terminal set                       & \{{\bf{Min}}, random(-1,1)\}                                           \\
Function set                       & \{+, -, ×, ÷ (protected)\}        
\\ 
The threshold of quality           & 1.0               \\
The number of nearest neighbors    & 3
\\ \hline
\end{tabular}
\end{table}
 
TABLE \ref{tab:parameter} shows parameter settings in our method. In EvoSampling, generating an instance is treated as a task. To generate $n$ (where $n=|{\bf{Maj}}|-|{\bf{Min}}|$) instances in $n$ tasks, $n$ GP processes are required, each handling a task and containing 30 individuals. Therefore, the total population size in EvoSampling equals $30 \times n$. The number of generations is set to 50, ensuring relatively enough evaluations. The tournament selection is used, with the tournament size of 3. 
The terminal set of EvoSampling includes {\bf{Min}} instances and a random number ranging from -1 to 1; the function set is \{+, -, ×, ÷\}, where ÷ is protected, returning 1 when the denominator is 0. The quality threshold for granule ball splitting is set to 1.0, and the number of nearest neighbors for GB undersampling is set to 3.

EvoSampling was implemented using the DEAP package \cite{fortin2012deap}, and the baseline sampling methods were implemented using the imbalanced-learn package \cite{lemaavztre2017imbalanced}. Adaboost \cite{ying2013advance}, gradient boosting decision trees (GBDT) \cite{ke2017lightgbm}, random forests (RF) \cite{breiman2001random}, and support vector machines (SVMs) \cite{chang2011libsvm}, use data rebalanced by our method and baseline methods to assess the effectiveness of these sampling techniques in addressing class imbalance. The four classification algorithms were implemented based on the scikit-learn package \cite{pedregosa2011scikit}, with default settings applied for simplicity.

\section{Results and Analysis}

On each dataset, EvoSampling has been independently executed 30 times, each with a different random seed. 
Using a sampling method, the rebalanced test sets are used by 4 different classifiers (including AdaBoost, GBDT, RF, and SVMs) to output their AUC \cite{fawcett2006introduction} and G\_Mean \cite{he2009learning} performance. 
We also conducted a Wilcoxon rank-sum test with a significance level of 0.05 to demonstrate the statistical significance of the performance differences between EvoSampling and a baseline method.

\subsection{Overall Results with Analysis}

\begin{table*}[]
\caption{Results of Sampling Methods with \textbf{Adaboost} on the test sets}
\centering
\label{tab:ADA_result}
\setlength{\tabcolsep}{3pt}
\renewcommand\arraystretch{1.4}
  \raggedright  
  \resizebox{\textwidth}{!}{%
\begin{tabular}{|c|cccccccccccccccccc|}
\hline
\multirow{3}{*}{Dataset} & \multicolumn{2}{c|}{\multirow{2}{*}{Initial}} & \multicolumn{2}{c|}{\multirow{2}{*}{SMOTE}} & \multicolumn{2}{c|}{\multirow{2}{*}{ADASYN}} & \multicolumn{2}{c|}{\multirow{2}{*}{Borderline-SMOTE1}} & \multicolumn{2}{c|}{\multirow{2}{*}{Borderline-SMOTE2}} & \multicolumn{2}{c|}{\multirow{2}{*}{SMOTE+ENN}} & \multicolumn{2}{c|}{\multirow{2}{*}{SMOTE+Tomek}} & \multicolumn{4}{c|}{\textbf{EvoSampling}} \\ \cline{16-19} 
 & \multicolumn{2}{c|}{} & \multicolumn{2}{c|}{} & \multicolumn{2}{c|}{} & \multicolumn{2}{c|}{} & \multicolumn{2}{c|}{} & \multicolumn{2}{c|}{} & \multicolumn{2}{c|}{} & \multicolumn{2}{c|}{Best} & \multicolumn{2}{c|}{Mean} \\ \cline{2-19} 
 & AUC & \multicolumn{1}{c|}{G\_Mean} & AUC & \multicolumn{1}{c|}{G\_Mean} & AUC & \multicolumn{1}{c|}{G\_Mean} & AUC & \multicolumn{1}{c|}{G\_Mean} & AUC & \multicolumn{1}{c|}{G\_Mean} & AUC & \multicolumn{1}{c|}{G\_Mean} & AUC & \multicolumn{1}{c|}{G\_Mean} & AUC & \multicolumn{1}{c|}{G\_Mean} & AUC & G\_Mean \\ \hline
D1 & \textbf{0.912 (-)} & \multicolumn{1}{c|}{\textbf{0.910 (-)}} & 0.870 (+) & \multicolumn{1}{c|}{\textbf{0.910 (-)}} & 0.887 (+) & \multicolumn{1}{c|}{0.886 (+)} & 0.879 (+) & \multicolumn{1}{c|}{0.878 (+)} & 0.887 (+) & \multicolumn{1}{c|}{0.886 (+)} & 0.895 (+) & \multicolumn{1}{c|}{0.894 (-)} & 0.870 (+) & \multicolumn{1}{c|}{0.870 (+)} & \textbf{0.912} & \multicolumn{1}{c|}{\textbf{0.910}} & 0.892 & 0.891 \\
D2 & 0.849 (+) & \multicolumn{1}{c|}{0.838 (+)} & 0.888 (-) & \multicolumn{1}{c|}{0.838 (+)} & 0.888 (-) & \multicolumn{1}{c|}{0.887 (+)} & 0.888 (-) & \multicolumn{1}{c|}{0.887 (+)} & 0.761 (+) & \multicolumn{1}{c|}{0.737 (+)} & 0.888 (-) & \multicolumn{1}{c|}{0.887 (+)} & 0.888 (-) & \multicolumn{1}{c|}{0.887 (+)} & \textbf{0.896} & \multicolumn{1}{c|}{\textbf{0.895}} & 0.888 & 0.888 \\
D3 & 0.899 (-) & \multicolumn{1}{c|}{0.896 (=)} & 0.824 (+) & \multicolumn{1}{c|}{0.896 (+)} & 0.899 (-) & \multicolumn{1}{c|}{0.896 (=)} & 0.908 (-) & \multicolumn{1}{c|}{0.905 (-)} & 0.899 (-) & \multicolumn{1}{c|}{0.896 (=)} & 0.667 (+) & \multicolumn{1}{c|}{0.577 (+)} & 0.824 (+) & \multicolumn{1}{c|}{0.809 (+)} & \textbf{0.973} & \multicolumn{1}{c|}{\textbf{0.973}} & 0.889 & 0.885 \\
D4 & 0.875 (=) & \multicolumn{1}{c|}{0.866 (=)} & 0.868 (=) & \multicolumn{1}{c|}{0.866 (+)} & 0.846 (+) & \multicolumn{1}{c|}{0.841 (+)} & 0.861 (+) & \multicolumn{1}{c|}{0.854 (+)} & 0.868 (=) & \multicolumn{1}{c|}{0.860 (+)} & 0.923 (-) & \multicolumn{1}{c|}{0.922 (-)} & 0.861 (+) & \multicolumn{1}{c|}{0.854 (+)} & \textbf{0.938} & \multicolumn{1}{c|}{\textbf{0.935}} & 0.878 & 0.870 \\
D5 & 0.833 (+) & \multicolumn{1}{c|}{0.816 (=)} & 0.833 (+) & \multicolumn{1}{c|}{0.816 (=)} & 0.833 (+) & \multicolumn{1}{c|}{0.816 (=)} & 0.833 (+) & \multicolumn{1}{c|}{0.816 (=)} & 0.833 (+) & \multicolumn{1}{c|}{0.816 (=)} & \textbf{1.000 (-)} & \multicolumn{1}{c|}{\textbf{1.000 (-)}} & 0.833 (+) & \multicolumn{1}{c|}{0.816 (=)} & \textbf{1.000} & \multicolumn{1}{c|}{\textbf{1.000}} & 0.821 & 0.765 \\
D6 & 0.500 (+) & \multicolumn{1}{c|}{0.000 (+)} & 0.582 (+) & \multicolumn{1}{c|}{0.000 (+)} & 0.565 (+) & \multicolumn{1}{c|}{0.431 (+)} & 0.565 (+) & \multicolumn{1}{c|}{0.431 (+)} & 0.682 (+) & \multicolumn{1}{c|}{0.621 (+)} & 0.604 (+) & \multicolumn{1}{c|}{0.568 (+)} & 0.674 (+) & \multicolumn{1}{c|}{0.616 (+)} & \textbf{0.930} & \multicolumn{1}{c|}{\textbf{0.927}} & 0.806 & 0.798 \\
D7 & 0.575 (+) & \multicolumn{1}{c|}{0.436 (+)} & 0.642 (+) & \multicolumn{1}{c|}{0.436 (+)} & 0.642 (+) & \multicolumn{1}{c|}{0.594 (+)} & 0.642 (+) & \multicolumn{1}{c|}{0.594 (+)} & 0.642 (+) & \multicolumn{1}{c|}{0.594 (+)} & 0.625 (+) & \multicolumn{1}{c|}{0.583 (+)} & 0.642 (+) & \multicolumn{1}{c|}{0.594 (+)} & \textbf{0.908} & \multicolumn{1}{c|}{\textbf{0.904}} & 0.758 & 0.753 \\
D8 & 0.859 (+) & \multicolumn{1}{c|}{0.852 (+)} & 0.864 (=) & \multicolumn{1}{c|}{0.852 (+)} & 0.870 (=) & \multicolumn{1}{c|}{0.861 (=)} & 0.842 (+) & \multicolumn{1}{c|}{0.837 (+)} & 0.875 (=) & \multicolumn{1}{c|}{0.866 (=)} & 0.864 (=) & \multicolumn{1}{c|}{0.857 (+)} & 0.864 (=) & \multicolumn{1}{c|}{0.857 (+)} & \textbf{0.932} & \multicolumn{1}{c|}{\textbf{0.930}} & 0.880 & 0.875 \\
D9 & 0.827(-) & \multicolumn{1}{c|}{0.811 (-)} & 0.744 (+) & \multicolumn{1}{c|}{0.811 (+)} & 0.744 (+) & \multicolumn{1}{c|}{0.703 (+)} & 0.744 (+) & \multicolumn{1}{c|}{0.703 (+)} & 0.744 (+) & \multicolumn{1}{c|}{0.703 (+)} & \textbf{0.917 (-)} & \multicolumn{1}{c|}{\textbf{0.913 (-)}} & 0.744 (+) & \multicolumn{1}{c|}{0.703 (+)} & 0.910 & \multicolumn{1}{c|}{0.907} & 0.836 & 0.822 \\
D10 & 0.552 (+) & \multicolumn{1}{c|}{0.332 (+)} & 0.553 (+) & \multicolumn{1}{c|}{0.332 (+)} & 0.549 (+) & \multicolumn{1}{c|}{0.441 (+)} & 0.497 (+) & \multicolumn{1}{c|}{0.313 (+)} & 0.497 (+) & \multicolumn{1}{c|}{0.313 (+)} & 0.656 (-) & \multicolumn{1}{c|}{0.621 (-)} & 0.565 (+) & \multicolumn{1}{c|}{0.449 (+)} & \textbf{0.740} & \multicolumn{1}{c|}{\textbf{0.737}} & 0.596 & 0.558 \\
D11 & 0.750 (+) & \multicolumn{1}{c|}{0.707 (+)} & 0.750 (+) & \multicolumn{1}{c|}{0.707 (+)} & 0.750 (+) & \multicolumn{1}{c|}{0.707 (+)} & 0.625 (+) & \multicolumn{1}{c|}{0.500 (+)} & 0.742 (+) & \multicolumn{1}{c|}{0.701 (+)} & 0.742 (+) & \multicolumn{1}{c|}{0.701 (+)} & 0.750 (+) & \multicolumn{1}{c|}{0.707 (+)} & \textbf{0.875} & \multicolumn{1}{c|}{\textbf{0.866}} & 0.830 & 0.823 \\
D12 & 0.546 (=) & \multicolumn{1}{c|}{0.330 (+)} & 0.510 (+) & \multicolumn{1}{c|}{0.330 (+)} & 0.541 (=) & \multicolumn{1}{c|}{0.437 (=)} & 0.518 (+) & \multicolumn{1}{c|}{0.321 (+)} & 0.523 (+) & \multicolumn{1}{c|}{0.322 (+)} & 0.511 (+) & \multicolumn{1}{c|}{0.421 (+)} & 0.510 (+) & \multicolumn{1}{c|}{0.318 (+)} & \textbf{0.596} & \multicolumn{1}{c|}{\textbf{0.535}} & 0.545 & 0.466 \\
D13 & 0.667 (=) & \multicolumn{1}{c|}{0.577 (=)} & 0.667 (=) & \multicolumn{1}{c|}{0.577 (=)} & 0.667 (=) & \multicolumn{1}{c|}{0.577 (=)} & 0.667 (=) & \multicolumn{1}{c|}{0.577 (=)} & 0.651 (+) & \multicolumn{1}{c|}{0.568 (=)} & 0.667 (=) & \multicolumn{1}{c|}{0.577 (=)} & 0.667 (=) & \multicolumn{1}{c|}{0.577 (=)} & \textbf{1.000} & \multicolumn{1}{c|}{\textbf{1.000}} & 0.745 & 0.646 \\
D14 & 0.493 (+) & \multicolumn{1}{c|}{0.000 (+)} & 0.438 (+) & \multicolumn{1}{c|}{0.000 (+)} & 0.451 (+) & \multicolumn{1}{c|}{0.000 (+)} & 0.451 (+) & \multicolumn{1}{c|}{0.000 (+)} & 0.493 (+) & \multicolumn{1}{c|}{0.000 (+)} & 0.424 (+) & \multicolumn{1}{c|}{0.000 (+)} & 0.438 (+) & \multicolumn{1}{c|}{0.000 (+)} & \textbf{0.736} & \multicolumn{1}{c|}{\textbf{0.697}} & 0.669 & 0.631 \\
D15 & 0.583 (+) & \multicolumn{1}{c|}{0.408 (+)} & 0.576 (+) & \multicolumn{1}{c|}{0.408 (+)} & 0.568 (+) & \multicolumn{1}{c|}{0.402 (+)} & 0.574 (+) & \multicolumn{1}{c|}{0.404 (+)} & 0.581 (+) & \multicolumn{1}{c|}{0.407 (+)} & 0.559 (+) & \multicolumn{1}{c|}{0.398 (+)} & 0.491 (+) & \multicolumn{1}{c|}{0.000 (+)} & \textbf{0.803} & \multicolumn{1}{c|}{\textbf{0.791}} & 0.650 & 0.583 \\
D16 & 0.560 (-) & \multicolumn{1}{c|}{0.353 (-)} & 0.518 (+) & \multicolumn{1}{c|}{0.353 (+)} & 0.517 (+) & \multicolumn{1}{c|}{0.337 (+)} & 0.487 (+) & \multicolumn{1}{c|}{0.000 (+)} & 0.560 (-) & \multicolumn{1}{c|}{0.353 (-)} & 0.554 (-) & \multicolumn{1}{c|}{0.463 (-)} & 0.516 (+) & \multicolumn{1}{c|}{0.337 (+)} & \textbf{0.598} & \multicolumn{1}{c|}{\textbf{0.486}} & 0.527 & 0.282 \\
D17 & 0.500 (+) & \multicolumn{1}{c|}{0.000 (+)} & 0.446 (+) & \multicolumn{1}{c|}{0.000 (+)} & 0.509 (+) & \multicolumn{1}{c|}{\textbf{0.438 (-)}} & 0.465 (+) & \multicolumn{1}{c|}{0.000 (+)} & 0.489 (+) & \multicolumn{1}{c|}{0.000 (+)} & 0.490 (+) & \multicolumn{1}{c|}{0.427 (-)} & 0.446 (+) & \multicolumn{1}{c|}{0.310 (+)} & \textbf{0.544} & \multicolumn{1}{c|}{0.347} & 0.520 & 0.317 \\
D18 & 0.500 (+) & \multicolumn{1}{c|}{0.000 (+)} & 0.463 (+) & \multicolumn{1}{c|}{0.000 (+)} & 0.468 (+) & \multicolumn{1}{c|}{0.000 (+)} & 0.480 (+) & \multicolumn{1}{c|}{0.000 (+)} & 0.662 (=) & \multicolumn{1}{c|}{0.575 (=)} & 0.618 (+) & \multicolumn{1}{c|}{0.548 (+)} & 0.463 (+) & \multicolumn{1}{c|}{0.000 (+)} & \textbf{0.956} & \multicolumn{1}{c|}{\textbf{0.955}} & 0.709 & 0.627 \\
D19 & 0.500 (+) & \multicolumn{1}{c|}{0.000 (+)} & 0.587 (=) & \multicolumn{1}{c|}{0.000 (-)} & 0.593 (-) & \multicolumn{1}{c|}{0.552 (-)} & 0.595 (-) & \multicolumn{1}{c|}{0.553 (-)} & 0.443 (+) & \multicolumn{1}{c|}{0.000 (+)} & 0.582 (=) & \multicolumn{1}{c|}{0.544 (-)} & 0.587 (=) & \multicolumn{1}{c|}{0.547 (-)} & \textbf{0.660} & \multicolumn{1}{c|}{\textbf{0.595}} & 0.552 & 0.361 \\
D20 & 0.500 (+) & \multicolumn{1}{c|}{0.000 (+)} & 0.496 (+) & \multicolumn{1}{c|}{0.000 (+)} & 0.488 (+) & \multicolumn{1}{c|}{0.394 (+)} & 0.500 (+) & \multicolumn{1}{c|}{0.000 (+)} & 0.500 (+) & \multicolumn{1}{c|}{0.000 (+)} & 0.477 (+) & \multicolumn{1}{c|}{0.388 (+)} & 0.496 (+) & \multicolumn{1}{c|}{0.398 (+)} & \textbf{0.628} & \multicolumn{1}{c|}{\textbf{0.585}} & 0.534 & 0.383 \\ \hline
\multicolumn{1}{|l|}{Overall} & \multicolumn{18}{c|}{200 +, 39 -, 41 =} \\ \hline
\end{tabular}
}
\end{table*}

\begin{table*}[]
\caption{Results of Sampling Methods with \textbf{GBDT} on the test sets}
\centering

\label{tab:GBDT_result}
\setlength{\tabcolsep}{3pt}
\renewcommand\arraystretch{1.4}

\begin{minipage}{1\textwidth}  
  \raggedright  
  \resizebox{\textwidth}{!}{%
\begin{tabular}{|c|llllllllllllllllll|}
\hline
\multirow{3}{*}{Dataset} & \multicolumn{2}{c|}{\multirow{2}{*}{Initial}} & \multicolumn{2}{c|}{\multirow{2}{*}{SMOTE}} & \multicolumn{2}{c|}{\multirow{2}{*}{ADASYN}} & \multicolumn{2}{c|}{\multirow{2}{*}{Borderline-SMOTE1}} & \multicolumn{2}{c|}{\multirow{2}{*}{Borderline-SMOTE2}} & \multicolumn{2}{c|}{\multirow{2}{*}{SMOTE+ENN}} & \multicolumn{2}{c|}{\multirow{2}{*}{SMOTE+Tomek}} & \multicolumn{4}{c|}{\textbf{EvoSampling}} \\ \cline{16-19} 
 & \multicolumn{2}{c|}{} & \multicolumn{2}{c|}{} & \multicolumn{2}{c|}{} & \multicolumn{2}{c|}{} & \multicolumn{2}{c|}{} & \multicolumn{2}{c|}{} & \multicolumn{2}{c|}{} & \multicolumn{2}{c|}{Best} & \multicolumn{2}{c|}{Mean} \\ \cline{2-19} 
 & \multicolumn{1}{c}{AUC} & \multicolumn{1}{c|}{G\_Mean} & \multicolumn{1}{c}{AUC} & \multicolumn{1}{c|}{G\_Mean} & \multicolumn{1}{c}{AUC} & \multicolumn{1}{c|}{G\_Mean} & \multicolumn{1}{c}{AUC} & \multicolumn{1}{c|}{G\_Mean} & \multicolumn{1}{c}{AUC} & \multicolumn{1}{c|}{G\_Mean} & \multicolumn{1}{c}{AUC} & \multicolumn{1}{c|}{G\_Mean} & \multicolumn{1}{c}{AUC} & \multicolumn{1}{c|}{G\_Mean} & \multicolumn{1}{c}{AUC} & \multicolumn{1}{c|}{G\_Mean} & \multicolumn{1}{c}{AUC} & \multicolumn{1}{c|}{G\_Mean} \\ \hline
D1 & \textbf{0.904 (-)} & \multicolumn{1}{l|}{\textbf{0.902 (-)}} & 0.664 (+) & \multicolumn{1}{l|}{0.621 (+)} & 0.681 (+) & \multicolumn{1}{l|}{0.632 (+)} & 0.879 (=) & \multicolumn{1}{l|}{0.878 (=)} & 0.887  (-) & \multicolumn{1}{l|}{0.886 (=)} & 0.895 (-) & \multicolumn{1}{l|}{0.894 (-)} & 0.664 (+) & \multicolumn{1}{l|}{0.621 (+)} & \textbf{0.904} & \multicolumn{1}{l|}{\textbf{0.902}} & 0.833 & 0.822 \\
D2 & 0.816  (+) & \multicolumn{1}{l|}{0.810 (+)} & 0.888  (-) & \multicolumn{1}{l|}{0.887  (=)} & 0.816  (+) & \multicolumn{1}{l|}{0.810 (+)} & 0.888  (-) & \multicolumn{1}{l|}{0.887  (=)} & 0.610 (+) & \multicolumn{1}{l|}{0.517  (+)} & 0.888  (-) & \multicolumn{1}{l|}{0.887  (=)} & 0.888  (-) & \multicolumn{1}{l|}{0.887  (=)} & \textbf{0.896} & \multicolumn{1}{l|}{\textbf{0.895}} & 0.811 & 0.797 \\
D3 & 0.908 (-) & \multicolumn{1}{l|}{0.905 (-)} & 0.723 (+) & \multicolumn{1}{l|}{0.688 (+)} & 0.890 (+) & \multicolumn{1}{l|}{0.888 (+)} & 0.741 (+) & \multicolumn{1}{l|}{0.701 (+)} & 0.890 (+) & \multicolumn{1}{l|}{0.888 (+)} & 0.649 (+) & \multicolumn{1}{l|}{0.567 (+)} & 0.723 (+) & \multicolumn{1}{l|}{0.688 (+)} & \textbf{0.973} & \multicolumn{1}{l|}{\textbf{0.973}} & 0.886 & 0.883 \\
D4 & 0.861 (+) & \multicolumn{1}{l|}{0.854 (+)} & 0.923 (-) & \multicolumn{1}{l|}{0.922 (-)} & 0.923 (-) & \multicolumn{1}{l|}{0.922 (-)} & \textbf{0.930 (-)} & \multicolumn{1}{l|}{\textbf{0.929 (-)}} & 0.923 (-) & \multicolumn{1}{l|}{0.922 (-)} & 0.923 (-) & \multicolumn{1}{l|}{0.922 (-)} & 0.916 (-) & \multicolumn{1}{l|}{0.915 (-)} & \textbf{0.930} & \multicolumn{1}{l|}{\textbf{0.929}} & 0.903 & 0.901 \\
D5 & \textbf{1.000 (-)} & \multicolumn{1}{l|}{\textbf{1.000 (-)}} & \textbf{1.000 (-)} & \multicolumn{1}{l|}{\textbf{1.000 (-)}} & \textbf{1.000 (-)} & \multicolumn{1}{l|}{\textbf{1.000 (-)}} & \textbf{1.000 (-)} & \multicolumn{1}{l|}{\textbf{1.000 (-)}} & \textbf{1.000 (-)} & \multicolumn{1}{l|}{\textbf{1.000 (-)}} & \textbf{1.000 (-)} & \multicolumn{1}{l|}{\textbf{1.000 (-)}} & \textbf{1.000 (-)} & \multicolumn{1}{l|}{\textbf{1.000 (-)}} & \textbf{1.000} & \multicolumn{1}{l|}{\textbf{1.000}} & 0.868 & 0.831 \\
D6 & 0.500 (+) & \multicolumn{1}{l|}{0.000 (+)} & 0.674 (+) & \multicolumn{1}{l|}{0.616 (+)} & 0.574 (+) & \multicolumn{1}{l|}{0.435 (+)} & 0.674 (+) & \multicolumn{1}{l|}{0.616 (+)} & 0.674 (+) & \multicolumn{1}{l|}{0.616 (+)} & 0.604 (+) & \multicolumn{1}{l|}{0.568 (+)} & 0.756 (=) & \multicolumn{1}{l|}{0.740 (=)} & \textbf{0.930} & \multicolumn{1}{l|}{\textbf{0.927}} & 0.766 & 0.751 \\
D7 & 0.558 (+) & \multicolumn{1}{l|}{0.428 (+)} & 0.750 (=) & \multicolumn{1}{l|}{0.735 (=)} & 0.683 (+) & \multicolumn{1}{l|}{0.678 (+)} & 0.633 (+) & \multicolumn{1}{l|}{0.589 (+)} & 0.467 (+) & \multicolumn{1}{l|}{0.000 (+)} & 0.708 (+) & \multicolumn{1}{l|}{0.700 (+)} & 0.750 (=) & \multicolumn{1}{l|}{0.735 (=)} & \textbf{0.833} & \multicolumn{1}{l|}{\textbf{0.833}} & 0.763 & 0.760 \\
D8 & 0.807 (+) & \multicolumn{1}{l|}{0.786 (+)} & 0.848 (+) & \multicolumn{1}{l|}{0.842  (+)} & 0.916 (=) & \multicolumn{1}{l|}{0.915 (=)} & 0.853 (+) & \multicolumn{1}{l|}{0.847 (+)} & 0.842  (+) & \multicolumn{1}{l|}{0.837 (+)} & 0.921 (-) & \multicolumn{1}{l|}{0.920 (-)} & 0.848 (+) & \multicolumn{1}{l|}{0.842  (+)} & \textbf{0.946} & \multicolumn{1}{l|}{\textbf{0.944}} & 0.912 & 0.911 \\
D9 & 0.744 (+) & \multicolumn{1}{l|}{0.703 (+)} & 0.750 (+) & \multicolumn{1}{l|}{0.707 (+)} & 0.744 (+) & \multicolumn{1}{l|}{0.703 (+)} & 0.660 (+) & \multicolumn{1}{l|}{0.574 (+)} & 0.744 (+) & \multicolumn{1}{l|}{0.703 (+)} & 0.744 (+) & \multicolumn{1}{l|}{0.703 (+)} & 0.750 (+) & \multicolumn{1}{l|}{0.707 (+)} & \textbf{0.833} & \multicolumn{1}{l|}{\textbf{0.816}} & 0.812 & 0.794 \\
D10 & 0.552 (+) & \multicolumn{1}{l|}{0.332 (+)} & 0.599 (=) & \multicolumn{1}{l|}{0.466 (+)} & 0.525 (+) & \multicolumn{1}{l|}{0.323 (+)} & 0.528 (+) & \multicolumn{1}{l|}{0.324 (+)} & 0.592 (=) & \multicolumn{1}{l|}{0.462 (+)} & 0.708 (-) & \multicolumn{1}{l|}{0.691 (-)} & 0.536 (+) & \multicolumn{1}{l|}{0.327 (+)} & \textbf{0.709} & \multicolumn{1}{l|}{\textbf{0.708}} & 0.593 & 0.555 \\
D11 & 0.750 (+) & \multicolumn{1}{l|}{0.707 (+)} & 0.875 (-) & \multicolumn{1}{l|}{0.866 (-)} & 0.875 (-) & \multicolumn{1}{l|}{0.866 (-)} & 0.875 (-) & \multicolumn{1}{l|}{0.866 (-)} & 0.875 (-) & \multicolumn{1}{l|}{0.866 (-)} & 0.867 (-) & \multicolumn{1}{l|}{0.859  (-)} & 0.875  (-) & \multicolumn{1}{l|}{0.866 (-)} & \textbf{0.967} & \multicolumn{1}{l|}{\textbf{0.967}} & 0.839 & 0.834 \\
D12 & 0.470  (+) & \multicolumn{1}{l|}{0.000 (+)} & 0.586  (-) & \multicolumn{1}{l|}{0.459  (-)} & 0.518  (+) & \multicolumn{1}{l|}{0.321  (+)} & 0.528  (=) & \multicolumn{1}{l|}{0.324  (+)} & 0.540  (-) & \multicolumn{1}{l|}{0.328  (+)} & 0.538  (-) & \multicolumn{1}{l|}{0.436  (-)} & 0.586 (-) & \multicolumn{1}{l|}{0.459 (-)} & \textbf{0.591} & \multicolumn{1}{l|}{\textbf{0.532}} & 0.532 & 0.442 \\
D13 & 0.984 (-) & \multicolumn{1}{l|}{0.984 (-)} & 0.833  (-) & \multicolumn{1}{l|}{0.816 (-)} & 0.833 (-) & \multicolumn{1}{l|}{0.816 (-)} & 0.833 (-) & \multicolumn{1}{l|}{0.816  (-)} & 0.651  (+) & \multicolumn{1}{l|}{0.568 (=)} & 0.833 (-) & \multicolumn{1}{l|}{0.816  (-)} & 0.833  (-) & \multicolumn{1}{l|}{0.816  (-)} & \textbf{1.000} & \multicolumn{1}{l|}{\textbf{1.000}} & 0.762 & 0.679 \\
D14 & 0.493  (+) & \multicolumn{1}{l|}{0.000 (+)} & 0.493  (+) & \multicolumn{1}{l|}{0.000 (+)} & 0.500 (+) & \multicolumn{1}{l|}{0.000 (+)} & 0.500 (+) & \multicolumn{1}{l|}{0.00 0 (+)} & 0.500 (+) & \multicolumn{1}{l|}{0.000  (+)} & 0.458 (+) & \multicolumn{1}{l|}{0.000 (+)} & 0.493  (+) & \multicolumn{1}{l|}{0.000  (+)} & \textbf{0.750} & \multicolumn{1}{l|}{\textbf{0.707}} & 0.644 & 0.444 \\
D15 & 0.667  (-) & \multicolumn{1}{l|}{0.577  (-)} & 0.494 (+) & \multicolumn{1}{l|}{0.000  (+)} & 0.494  (+) & \multicolumn{1}{l|}{0.000  (+)} & 0.496  (+)  & \multicolumn{1}{l|}{0.000  (+)} & 0.665 (-) & \multicolumn{1}{l|}{0.576 (-)} & 0.646 (-) & \multicolumn{1}{l|}{0.565 (-)} & 0.492 (+) & \multicolumn{1}{l|}{0.000 (+)} & \textbf{0.710} & \multicolumn{1}{l|}{\textbf{0.678}} & 0.627 & 0.552 \\
D16 & 0.498 (=) & \multicolumn{1}{l|}{0.000 (+)} & 0.494 (=) & \multicolumn{1}{l|}{0.000 (+)} & 0.490 (=) & \multicolumn{1}{l|}{0.000 (+)} & 0.492 (=) & \multicolumn{1}{l|}{0.000 (+)} & 0.499 (=) & \multicolumn{1}{l|}{0.000 (+)} & 0.471 (+) & \multicolumn{1}{l|}{0.000 (+)} & 0.494 (=) & \multicolumn{1}{l|}{0.000 (+)} & \textbf{0.713} & \multicolumn{1}{l|}{\textbf{0.681}} & 0.521 & 0.202 \\
D17 & 0.499 (+) & \multicolumn{1}{l|}{0.000 (+)} & 0.813 (-) & \multicolumn{1}{l|}{0.791 (-)} & 0.745 (-) & \multicolumn{1}{l|}{0.704 (-)} & 0.498 (+) & \multicolumn{1}{l|}{0.000 (+)} & 0.500 (+) & \multicolumn{1}{l|}{0.000 (+)} & 0.743 (-) & \multicolumn{1}{l|}{0.702 (-)} & 0.813 (-) & \multicolumn{1}{l|}{0.791 (-)} & \textbf{1.000} & \multicolumn{1}{l|}{\textbf{1.000}} & 0.668 & 0.485 \\
D18 & 0.500 (+) & \multicolumn{1}{l|}{0.000 (+)} & 0.488 (+) & \multicolumn{1}{l|}{0.000 (+)} & 0.483 (+) & \multicolumn{1}{l|}{0.000 (+)} & 0.652 (-) & \multicolumn{1}{l|}{0.569 (-)} & 0.500 (+) & \multicolumn{1}{l|}{0.000 (+)} & 0.637 (=) & \multicolumn{1}{l|}{0.560 (=)} & 0.490 (+) & \multicolumn{1}{l|}{0.000 (+)} & \textbf{0.804} & \multicolumn{1}{l|}{\textbf{0.792}} & 0.650 & 0.574 \\
D19 & 0.500 (+) & \multicolumn{1}{l|}{0.000 (+)} & 0.554 (-) & \multicolumn{1}{l|}{0.351 (-)} & 0.613 (-) & \multicolumn{1}{l|}{\textbf{0.494 (-)}} & 0.558 (-) & \multicolumn{1}{l|}{0.352 (-)} & 0.497 (+) & \multicolumn{1}{l|}{0.000 (+)} & 0.548 (+) & \multicolumn{1}{l|}{0.348 (+)} & 0.554 (-) & \multicolumn{1}{l|}{0.351 (-)} & \textbf{0.609} & \multicolumn{1}{l|}{0.492} & 0.558 & 0.365 \\
D20 & 0.700 (+) & \multicolumn{1}{l|}{0.632 (+)} & 0.800 (=) & \multicolumn{1}{l|}{0.775 (+)} & 1.000 (-) & \multicolumn{1}{l|}{\textbf{1.000 (-)}} & 0.700 (+) & \multicolumn{1}{l|}{0.632 (+)} & 0.700 (+) & \multicolumn{1}{l|}{0.632 (+)} & 0.800 (=) & \multicolumn{1}{l|}{0.775 (+)} & 0.800 (=) & \multicolumn{1}{l|}{0.775 (+)} & \textbf{1.000} & \multicolumn{1}{l|}{\textbf{1.000}} & 0.846 & 0.823 \\ \hline
Overall & \multicolumn{18}{c|}{152 +, 98 -, 30 =} \\ \hline
\end{tabular}
}
\end{minipage}
\end{table*}

\begin{table*}[]
\caption{Results of Sampling Methods with \textbf{RF} on the test sets}
\centering
\label{tab:RF_result}
\setlength{\tabcolsep}{3pt}
\renewcommand\arraystretch{1.4}

\begin{minipage}{1\textwidth}  
  \raggedright  
  \resizebox{\textwidth}{!}{%
\begin{tabular}{|c|llllllllllllllllll|}
\hline
\multirow{3}{*}{Dataset} & \multicolumn{2}{c|}{\multirow{2}{*}{Initial}} & \multicolumn{2}{c|}{\multirow{2}{*}{SMOTE}} & \multicolumn{2}{c|}{\multirow{2}{*}{ADASYN}} & \multicolumn{2}{c|}{\multirow{2}{*}{Borderline-SMOTE1}} & \multicolumn{2}{c|}{\multirow{2}{*}{Borderline-SMOTE2}} & \multicolumn{2}{c|}{\multirow{2}{*}{SMOTE+ENN}} & \multicolumn{2}{c|}{\multirow{2}{*}{SMOTE+Tomek}} & \multicolumn{4}{c|}{\textbf{\textbf{EvoSampling}}} \\ \cline{16-19} 
 & \multicolumn{2}{c|}{} & \multicolumn{2}{c|}{} & \multicolumn{2}{c|}{} & \multicolumn{2}{c|}{} & \multicolumn{2}{c|}{} & \multicolumn{2}{c|}{} & \multicolumn{2}{c|}{} & \multicolumn{2}{c|}{Best} & \multicolumn{2}{c|}{Mean} \\ \cline{2-19} 
 & \multicolumn{1}{c}{AUC} & \multicolumn{1}{c|}{G\_Mean} & \multicolumn{1}{c}{AUC} & \multicolumn{1}{c|}{G\_Mean} & \multicolumn{1}{c}{AUC} & \multicolumn{1}{c|}{G\_Mean} & \multicolumn{1}{c}{AUC} & \multicolumn{1}{c|}{G\_Mean} & \multicolumn{1}{c}{AUC} & \multicolumn{1}{c|}{G\_Mean} & \multicolumn{1}{c}{AUC} & \multicolumn{1}{c|}{G\_Mean} & \multicolumn{1}{c}{AUC} & \multicolumn{1}{c|}{G\_Mean} & \multicolumn{1}{c}{AUC} & \multicolumn{1}{c|}{G\_Mean} & \multicolumn{1}{c}{AUC} & \multicolumn{1}{c|}{G\_Mean} \\ \hline
D1 & \textbf{0.920 (-)} & \multicolumn{1}{l|}{\textbf{0.918 (-)}} & 0.845 (+) & \multicolumn{1}{l|}{0.845 (+)} & 0.887  (+) & \multicolumn{1}{l|}{0.886 (+)} & 0.895 (+) & \multicolumn{1}{l|}{0.894 (=)} & 0.832 (+) & \multicolumn{1}{l|}{0.824 (+)} & 0.904 (-) & \multicolumn{1}{l|}{0.902 (=)} & 0.845 (+) & \multicolumn{1}{l|}{0.845 (+)} & 0.912 & \multicolumn{1}{l|}{0.910} & 0.887 & 0.884 \\
D2 & \textbf{0.929 (-)} & \multicolumn{1}{l|}{\textbf{0.926 (-)}} & 0.888  (-) & \multicolumn{1}{l|}{0.887  (+)} & 0.888  (-) & \multicolumn{1}{l|}{0.887  (+)} & 0.816  (+) & \multicolumn{1}{l|}{0.810 (+)} & 0.816 (+) & \multicolumn{1}{l|}{0.810 (+)} & 0.888 (-) & \multicolumn{1}{l|}{0.887  (+)} & 0.888 (-) & \multicolumn{1}{l|}{0.887  (+)} & 0.896 & \multicolumn{1}{l|}{0.895} & 0.882 & 0.881 \\
D3 & 0.750 (+) & \multicolumn{1}{l|}{0.707 (+)} & 0.750 (+) & \multicolumn{1}{l|}{0.707 (+)} & 0.908 (-) & \multicolumn{1}{l|}{0.905 (-)} & 0.750 (+) & \multicolumn{1}{l|}{0.707 (+)} & 0.750 (+) & \multicolumn{1}{l|}{0.707 (+)} & 0.833  (+) & \multicolumn{1}{l|}{0.816 (+)} & 0.750 (+) & \multicolumn{1}{l|}{0.707 (+)} & \textbf{0.991} & \multicolumn{1}{l|}{\textbf{0.991}} & 0.898 & 0.893 \\
D4 & 0.688 (+) & \multicolumn{1}{l|}{0.612 (+)} & 0.909 (+) & \multicolumn{1}{l|}{0.908 (+)} & 0.846 (+) & \multicolumn{1}{l|}{0.841 (+)} & 0.930 (-) & \multicolumn{1}{l|}{0.929 (+)} & 0.868 (+) & \multicolumn{1}{l|}{0.860 (+)} & 0.909 (+) & \multicolumn{1}{l|}{0.908 (+)} & 0.909 (+) & \multicolumn{1}{l|}{0.908 (+)} & \textbf{0.938} & \multicolumn{1}{l|}{\textbf{0.935}} & 0.919 & 0.916 \\
D5 & 0.500 (+) & \multicolumn{1}{l|}{0.000 (+)} & \textbf{1.000 (-)} & \multicolumn{1}{l|}{\textbf{1.000 (-)}} & 0.833  (+) & \multicolumn{1}{l|}{0.816 (+)} & \textbf{1.000 (-)} & \multicolumn{1}{l|}{\textbf{1.000 (-)}} & \textbf{1.000 (-)} & \multicolumn{1}{l|}{\textbf{1.000 (-)}} & 0.833  (+) & \multicolumn{1}{l|}{0.816 (+)} & \textbf{1.000 (-)} & \multicolumn{1}{l|}{\textbf{1.000 (-)}} & \textbf{1.000} & \multicolumn{1}{l|}{\textbf{1.000}} & 0.888 & 0.878 \\
D6 & 0.500 (+) & \multicolumn{1}{l|}{0.000 (+)} & 0.647 (+) & \multicolumn{1}{l|}{0.598 (+)} & 0.574 (+) & \multicolumn{1}{l|}{0.435 (+)} & 0.656 (+) & \multicolumn{1}{l|}{0.604 (+)} & 0.656 (+) & \multicolumn{1}{l|}{0.604 (+)} & 0.686 (+) & \multicolumn{1}{l|}{0.681 (+)} & 0.647 (+) & \multicolumn{1}{l|}{0.598 (+)} & \textbf{0.912} & \multicolumn{1}{l|}{\textbf{0.908}} & 0.787 & 0.778 \\
D7 & 0.500 (+) & \multicolumn{1}{l|}{0.000 (+)} & 0.633 (+) & \multicolumn{1}{l|}{0.589 (+)} & 0.742 (+) & \multicolumn{1}{l|}{0.728 (+)} & 0.742 (+) & \multicolumn{1}{l|}{0.728 (+)} & 0.567 (+) & \multicolumn{1}{l|}{0.432 (+)} & 0.633 (+) & \multicolumn{1}{l|}{0.589 (+)} & 0.633 (+) & \multicolumn{1}{l|}{0.589 (+)} & \textbf{0.808} & \multicolumn{1}{l|}{\textbf{0.808}} & 0.760 & 0.757 \\
D8 & 0.870 (+) & \multicolumn{1}{l|}{0.861 (+)} & 0.864 (+) & \multicolumn{1}{l|}{0.857 (+)} & 0.864 (+) & \multicolumn{1}{l|}{0.857 (+)} & 0.859 (+) & \multicolumn{1}{l|}{0.852 (+)} & 0.870 (+) & \multicolumn{1}{l|}{0.861 (+)} & 0.864 (+) & \multicolumn{1}{l|}{0.857 (+)} & 0.864 (+) & \multicolumn{1}{l|}{0.857 (+)} & \textbf{0.932} & \multicolumn{1}{l|}{\textbf{0.930}} & 0.921 & 0.919 \\
D9 & 0.667 (+) & \multicolumn{1}{l|}{0.577 (+)} & 0.744 (+) & \multicolumn{1}{l|}{0.703 (+)} & 0.744 (+) & \multicolumn{1}{l|}{0.703 (+)} & 0.660 (+) & \multicolumn{1}{l|}{0.574 (+)} & 0.744 (+) & \multicolumn{1}{l|}{0.703 (+)} & 0.750 (+) & \multicolumn{1}{l|}{0.707 (+)} & 0.744 (+) & \multicolumn{1}{l|}{0.703 (+)} & \textbf{0.827} & \multicolumn{1}{l|}{\textbf{0.811}} & 0.824 & 0.807 \\
D10 & 0.552 (+) & \multicolumn{1}{l|}{0.332 (+)} & 0.540 (+) & \multicolumn{1}{l|}{0.328 (+)} & 0.647 (-) & \multicolumn{1}{l|}{0.566 (+)} & 0.488 (+) & \multicolumn{1}{l|}{0.000 (+)} & 0.536 (+) & \multicolumn{1}{l|}{0.327 (+)} & 0.561 (+) & \multicolumn{1}{l|}{0.447 (+)} & 0.544 (+) & \multicolumn{1}{l|}{0.329 (+)} & \textbf{0.709} & \multicolumn{1}{l|}{\textbf{0.708}} & 0.610 & 0.584 \\
D11 & 0.725 (+) & \multicolumn{1}{l|}{0.689 (+)} & 0.867 (-) & \multicolumn{1}{l|}{0.859 (-)} & 0.867 (-) & \multicolumn{1}{l|}{0.859 (-)} & 0.875 (-) & \multicolumn{1}{l|}{0.866 (-)} & 0.867 (-) & \multicolumn{1}{l|}{0.859 (-)} & 0.850 (-) & \multicolumn{1}{l|}{0.844 (-)} & 0.867 (-) & \multicolumn{1}{l|}{0.859 (-)} & \textbf{0.943} & \multicolumn{1}{l|}{\textbf{0.941}} & 0.827 & 0.822 \\
D12 & 0.500 (+) & \multicolumn{1}{l|}{0.000 (+)} & 0.525 (=) & \multicolumn{1}{l|}{0.323 (+)} & 0.541 (=) & \multicolumn{1}{l|}{0.437 (=)} & 0.470 (+) & \multicolumn{1}{l|}{0.000 (+)} & 0.538 (=) & \multicolumn{1}{l|}{0.327 (+)} & 0.533 (=) & \multicolumn{1}{l|}{0.433 (=)} & 0.525 (=) & \multicolumn{1}{l|}{0.323 (+)} & \textbf{0.589} & \multicolumn{1}{l|}{\textbf{0.530}} & 0.522 & 0.449 \\
D13 & 0.500 (+) & \multicolumn{1}{l|}{0.000 (+)} & \textbf{1.000 (-)} & \multicolumn{1}{l|}{\textbf{1.000 (-)}} & 0.833  (-) & \multicolumn{1}{l|}{0.816 (-)} & \textbf{1.000 (-)} & \multicolumn{1}{l|}{\textbf{1.000 (-)}} & 0.833  (-) & \multicolumn{1}{l|}{0.816  (-)} & 0.833  (-) & \multicolumn{1}{l|}{0.816 (-)} & \textbf{1.000 (-)} & \multicolumn{1}{l|}{\textbf{1.000 (-)}} & \textbf{1.000} & \multicolumn{1}{l|}{\textbf{1.000}} & 0.797 & 0.768 \\
D14 & 0.500 (+) & \multicolumn{1}{l|}{0.000 (+)} & 0.500 (+) & \multicolumn{1}{l|}{0.000 (+)} & 0.500 (+) & \multicolumn{1}{l|}{0.000 (+)} & 0.500 (+) & \multicolumn{1}{l|}{0.000 (+)} & 0.500 (+) & \multicolumn{1}{l|}{0.000 (+)} & 0.479 (+) & \multicolumn{1}{l|}{0.000 (+)} & 0.500 (+) & \multicolumn{1}{l|}{0.000 (+)} & \textbf{0.750} & \multicolumn{1}{l|}{\textbf{0.707}} & 0.631 & 0.417 \\
D15 & 0.583 (+) & \multicolumn{1}{l|}{0.408 (+)} & 0.496 (+) & \multicolumn{1}{l|}{0.000 (+)} & 0.496 (+) & \multicolumn{1}{l|}{0.000 (+)} & 0.498 (+) & \multicolumn{1}{l|}{0.000 (+)} & 0.498 (+) & \multicolumn{1}{l|}{0.000 (+)} & 0.494 (+) & \multicolumn{1}{l|}{0.000 (+)} & 0.496 (+) & \multicolumn{1}{l|}{0.000 (+)} & \textbf{0.780} & \multicolumn{1}{l|}{\textbf{0.772}} & 0.647 & 0.597 \\
D16 & 0.499 (+) & \multicolumn{1}{l|}{0.000 (+)} & 0.554 (=) & \multicolumn{1}{l|}{0.351 (=)} & 0.494 (+) & \multicolumn{1}{l|}{0.000 (+)} & 0.497 (+) & \multicolumn{1}{l|}{0.000 (+)} & 0.499 (+) & \multicolumn{1}{l|}{0.000 (+)} & 0.476 (+) & \multicolumn{1}{l|}{0.000 (+)} & 0.554 (=) & \multicolumn{1}{l|}{0.351 (=)} & \textbf{0.716} & \multicolumn{1}{l|}{\textbf{0.682}} & 0.569 & 0.371 \\
D17 & 0.500 (+) & \multicolumn{1}{l|}{0.000 (+)} & 0.748 (=) & \multicolumn{1}{l|}{0.705 (=)} & 0.678 (=) & \multicolumn{1}{l|}{0.607 (=)} & 0.498 (+) & \multicolumn{1}{l|}{0.000 (+)} & 0.500 (+) & \multicolumn{1}{l|}{0.000 (+)} & 0.809 (-) & \multicolumn{1}{l|}{0.788 (-)} & 0.748 (=) & \multicolumn{1}{l|}{0.705 (=)} & \textbf{0.938} & \multicolumn{1}{l|}{\textbf{0.935}} & 0.689 & 0.588 \\
D18 & 0.500 (+) & \multicolumn{1}{l|}{0.000 (+)} & 0.490 (+) & \multicolumn{1}{l|}{0.000 (+)} & 0.488 (+) & \multicolumn{1}{l|}{0.000 (+)} & 0.662 (-) & \multicolumn{1}{l|}{0.575 (-)} & 0.500 (+) & \multicolumn{1}{l|}{0.000 (+)} & 0.628 (=) & \multicolumn{1}{l|}{0.554 (=)} & 0.485 (+) & \multicolumn{1}{l|}{0.000 (+)} & \textbf{0.802} & \multicolumn{1}{l|}{\textbf{0.790}} & 0.668 & 0.613 \\
D19 & 0.500 (-) & \multicolumn{1}{l|}{0.000 (+)} & 0.610 (-) & \multicolumn{1}{l|}{0.493 (-)} & 0.546 (-) & \multicolumn{1}{l|}{0.348 (-)} & \textbf{0.616 (-)} & \multicolumn{1}{l|}{\textbf{0.496 (-)}} & 0.500 (-) & \multicolumn{1}{l|}{0.000 (+)} & 0.545 (-) & \multicolumn{1}{l|}{0.347 (-)} & 0.610 (-) & \multicolumn{1}{l|}{0.493 (-)} & 0.547 & \multicolumn{1}{l|}{0.348} & 0.489 & 0.076 \\
D20 & 0.600 (+) & \multicolumn{1}{l|}{0.447 (+)} & \textbf{0.800 (-)} & \multicolumn{1}{l|}{\textbf{0.775 (=)}} & 0.700 (+) & \multicolumn{1}{l|}{0.632 (=)} & 0.600 (+) & \multicolumn{1}{l|}{0.447 (+)} & 0.600 (+) & \multicolumn{1}{l|}{0.447 (+)} & \textbf{0.800 (-)} & \multicolumn{1}{l|}{\textbf{0.775 (=)}} & \textbf{0.800 (-)} & \multicolumn{1}{l|}{\textbf{0.775 (=)}} & \textbf{0.800} & \multicolumn{1}{l|}{\textbf{0.775}} & 0.758 & 0.716 \\ \hline
Overall & \multicolumn{18}{c|}{191 +, 64 -, 25 =} \\ \hline
\end{tabular}
}
\end{minipage}
\end{table*}

\begin{table*}[h]
\caption{Results of Sampling Methods with \textbf{SVMs} on the test sets}
\centering

\label{tab:SVM_result}
\setlength{\tabcolsep}{3pt}
\renewcommand\arraystretch{1.4}

\begin{minipage}{1\textwidth}  
  \raggedright  
  \resizebox{\textwidth}{!}{%
\begin{tabular}{|c|llllllllllllllllll|}
\hline
\multirow{3}{*}{Dataset} & \multicolumn{2}{c|}{\multirow{2}{*}{Initial}} & \multicolumn{2}{c|}{\multirow{2}{*}{SMOTE}} & \multicolumn{2}{c|}{\multirow{2}{*}{ADASYN}} & \multicolumn{2}{c|}{\multirow{2}{*}{Borderline-SMOTE1}} & \multicolumn{2}{c|}{\multirow{2}{*}{Borderline-SMOTE2}} & \multicolumn{2}{c|}{\multirow{2}{*}{SMOTE+ENN}} & \multicolumn{2}{c|}{\multirow{2}{*}{SMOTE+Tomek}} & \multicolumn{4}{c|}{\textbf{EvoSampling}} \\ \cline{16-19} 
 & \multicolumn{2}{c|}{} & \multicolumn{2}{c|}{} & \multicolumn{2}{c|}{} & \multicolumn{2}{c|}{} & \multicolumn{2}{c|}{} & \multicolumn{2}{c|}{} & \multicolumn{2}{c|}{} & \multicolumn{2}{c|}{Best} & \multicolumn{2}{c|}{Mean} \\ \cline{2-19} 
 & \multicolumn{1}{c}{AUC} & \multicolumn{1}{c|}{G\_Mean} & \multicolumn{1}{c}{AUC} & \multicolumn{1}{c|}{G\_Mean} & \multicolumn{1}{c}{AUC} & \multicolumn{1}{c|}{G\_Mean} & \multicolumn{1}{c}{AUC} & \multicolumn{1}{c|}{G\_Mean} & \multicolumn{1}{c}{AUC} & \multicolumn{1}{c|}{G\_Mean} & \multicolumn{1}{c}{AUC} & \multicolumn{1}{c|}{G\_Mean} & \multicolumn{1}{c}{AUC} & \multicolumn{1}{c|}{G\_Mean} & \multicolumn{1}{c}{AUC} & \multicolumn{1}{c|}{G\_Mean} & \multicolumn{1}{c}{AUC} & \multicolumn{1}{c|}{G\_Mean} \\ \hline
D1 & 0.500 (+) & \multicolumn{1}{l|}{0.000 (+)} & 0.500 (+) & \multicolumn{1}{l|}{0.000 (+)} & 0.500 (+) & \multicolumn{1}{l|}{0.000 (+)} & 0.500 (+) & \multicolumn{1}{l|}{0.000 (+)} & 0.500 (+) & \multicolumn{1}{l|}{0.000 (+)} & 0.500 (+) & \multicolumn{1}{l|}{0.000 (+)} & 0.500 (+) & \multicolumn{1}{l|}{0.000 (+)} & \textbf{0.920} & \multicolumn{1}{l|}{\textbf{0.918}} & 0.778 & 0.769 \\
D2 & 0.871 (-) & \multicolumn{1}{l|}{0.871 (-)} & 0.500 (+) & \multicolumn{1}{l|}{0.000 (+)} & 0.806 (+) & \multicolumn{1}{l|}{0.804 (=)} & 0.500 (+) & \multicolumn{1}{l|}{0.000 (+)} & 0.500 (+) & \multicolumn{1}{l|}{0.000 (+)} & 0.765 (+) & \multicolumn{1}{l|}{0.759 (+)} & 0.500 (+) & \multicolumn{1}{l|}{0.000 (+)} & \textbf{0.879} & \multicolumn{1}{l|}{\textbf{0.879}} & 0.821 & 0.819 \\
D3 & 0.500 (+) & \multicolumn{1}{l|}{0.000 (+)} & 0.500 (+) & \multicolumn{1}{l|}{0.000 (+)} & 0.500 (+) & \multicolumn{1}{l|}{0.000 (+)} & 0.500 (+) & \multicolumn{1}{l|}{0.000 (+)} & 0.500 (+) & \multicolumn{1}{l|}{0.000 (+)} & 0.833  (=) & \multicolumn{1}{l|}{0.816 (=)} & 0.500 (+) & \multicolumn{1}{l|}{0.000 (+)} & \textbf{0.946} & \multicolumn{1}{l|}{\textbf{0.945}} & 0.750 & 0.630 \\
D4 & 0.500 (+) & \multicolumn{1}{l|}{0.000 (+)} & 0.500 (+) & \multicolumn{1}{l|}{0.000 (+)} & 0.500 (+) & \multicolumn{1}{l|}{0.000 (+)} & 0.500 (+) & \multicolumn{1}{l|}{0.000 (+)} & 0.500 (+) & \multicolumn{1}{l|}{0.000 (+)} & 0.630 (-) & \multicolumn{1}{l|}{0.576 (-)} & 0.500 (+) & \multicolumn{1}{l|}{0.000 (+)} & \textbf{0.829} & \multicolumn{1}{l|}{\textbf{0.811}} & 0.580 & 0.328 \\
D5 & \textbf{1.000 (-)} & \multicolumn{1}{l|}{\textbf{1.000 (-)}} & \textbf{1.000 (-)} & \multicolumn{1}{l|}{\textbf{1.000 (-)}} & 0.817 (-) & \multicolumn{1}{l|}{0.796 (-)} & 0.967 (-) & \multicolumn{1}{l|}{0.966 (-)} & 0.983 (-) & \multicolumn{1}{l|}{0.983 (-)} & 0.750 (=) & \multicolumn{1}{l|}{0.707 (=)} & \textbf{1.000 (-)} & \multicolumn{1}{l|}{\textbf{1.000 (-)}} & \textbf{1.000} & \multicolumn{1}{l|}{\textbf{1.000}} & 0.713 & 0.569 \\
D6 & 0.500 (+) & \multicolumn{1}{l|}{0.000 (+)} & 0.682 (-) & \multicolumn{1}{l|}{0.621 (-)} & 0.465 (+) & \multicolumn{1}{l|}{0.000 (+)} & 0.574 (=) & \multicolumn{1}{l|}{0.435 (=)} & 0.812 (-) & \multicolumn{1}{l|}{0.812 (-)} & 0.895 (-) & \multicolumn{1}{l|}{0.889 (-)} & 0.482 (+) & \multicolumn{1}{l|}{0.000 (+)} & \textbf{0.860} & \multicolumn{1}{l|}{\textbf{0.848}} & 0.583 & 0.359 \\
D7 & 0.500 (=) & \multicolumn{1}{l|}{0.000 (+)} & 0.500 (=) & \multicolumn{1}{l|}{0.000 (+)} & 0.492 (+) & \multicolumn{1}{l|}{0.000 (+)} & 0.492 (+) & \multicolumn{1}{l|}{0.000 (+)} & 0.500 (=) & \multicolumn{1}{l|}{0.000 (+)} & 0.500 (=) & \multicolumn{1}{l|}{0.000 (+)} & 0.500 (=) & \multicolumn{1}{l|}{0.000 (+)} & \textbf{0.750} & \multicolumn{1}{l|}{\textbf{0.735}} & 0.532 & 0.173 \\
D8 & 0.927 (-) & \multicolumn{1}{l|}{0.925 (-)} & \textbf{0.932 (-)} & \multicolumn{1}{l|}{\textbf{0.930 (-)}} & 0.916 (-) & \multicolumn{1}{l|}{0.915 (-)} & \textbf{0.932 (-)} & \multicolumn{1}{l|}{\textbf{0.930 (-)}} & \textbf{0.932 (-)} & \multicolumn{1}{l|}{\textbf{0.930 (-)}} & 0.864 (+) & \multicolumn{1}{l|}{0.857 (+)} & \textbf{0.932 (-)} & \multicolumn{1}{l|}{\textbf{0.930 (-)}} & \textbf{0.932} & \multicolumn{1}{l|}{\textbf{0.930}} & 0.890 & 0.888 \\
D9 & 0.500 (+) & \multicolumn{1}{l|}{0.000 (+)} & 0.833  (-) & \multicolumn{1}{l|}{0.816 (-)} & 0.827 (-) & \multicolumn{1}{l|}{0.811 (-)} & 0.833  (-) & \multicolumn{1}{l|}{0.816 (-)} & 0.833  (-) & \multicolumn{1}{l|}{0.816 (-)} & 0.833  (-) & \multicolumn{1}{l|}{0.816 (-)} & 0.833  (-) & \multicolumn{1}{l|}{0.816 (-)} & \textbf{0.891} & \multicolumn{1}{l|}{\textbf{0.889}} & 0.806 & 0.791 \\
D10 & 0.584 (+) & \multicolumn{1}{l|}{0.458 (+)} & 0.473 (+) & \multicolumn{1}{l|}{0.000 (+)} & 0.469 (+) & \multicolumn{1}{l|}{0.000 (+)} & 0.450 (+) & \multicolumn{1}{l|}{0.000 (+)} & 0.473 (+) & \multicolumn{1}{l|}{0.000 (+)} & 0.446 (+) & \multicolumn{1}{l|}{0.000 (+)} & 0.473 (+) & \multicolumn{1}{l|}{0.000 (+)} & \textbf{0.779} & \multicolumn{1}{l|}{\textbf{0.747}} & 0.653 & 0.598 \\
D11 & 0.625 (-) & \multicolumn{1}{l|}{0.500 (=)} & 0.625 (-) & \multicolumn{1}{l|}{0.500 (=)} & 0.625 (-) & \multicolumn{1}{l|}{0.500 (=)} & 0.625 (-) & \multicolumn{1}{l|}{0.500 (=)} & 0.500 (+) & \multicolumn{1}{l|}{0.000 (+)} & 0.750 (-) & \multicolumn{1}{l|}{0.707 (-)} & 0.625 (-) & \multicolumn{1}{l|}{0.500 (=)} & \textbf{0.975} & \multicolumn{1}{l|}{\textbf{0.975}} & 0.628 & 0.353 \\
D12 & 0.500 (+) & \multicolumn{1}{l|}{0.000 (+)} & 0.495 (+) & \multicolumn{1}{l|}{0.000 (+)} & 0.495 (+) & \multicolumn{1}{l|}{0.000 (+)} & 0.495 (+) & \multicolumn{1}{l|}{0.000 (+)} & 0.487 (+) & \multicolumn{1}{l|}{0.000 (+)} & 0.543 (+) & \multicolumn{1}{l|}{0.329 (+)} & 0.495 (+) & \multicolumn{1}{l|}{0.000 (+)} & \textbf{0.680} & \multicolumn{1}{l|}{\textbf{0.668}} & 0.585 & 0.554 \\
D13 & \textbf{0.667 (-)} & \multicolumn{1}{l|}{\textbf{0.577 (-)}} & 0.500 (=) & \multicolumn{1}{l|}{0.000 (+)} & 0.500 (=) & \multicolumn{1}{l|}{0.000 (+)} & 0.500 (=) & \multicolumn{1}{l|}{0.000 (+)} & 0.500 (=) & \multicolumn{1}{l|}{0.000 (+)} & 0.500 (=) & \multicolumn{1}{l|}{0.000 (+)} & 0.500 (=) & \multicolumn{1}{l|}{0.000 (+)} & 0.659 & \multicolumn{1}{l|}{0.573} & 0.522 & 0.094 \\
D14 & 0.500 (+) & \multicolumn{1}{l|}{0.000 (+)} & \textbf{0.750 (-)} & \multicolumn{1}{l|}{\textbf{0.707 (-)}} & 0.500 (+) & \multicolumn{1}{l|}{0.000 (+)} & 0.500 (+) & \multicolumn{1}{l|}{0.000 (+)} & 0.500 (+) & \multicolumn{1}{l|}{0.000 (+)} & \textbf{0.750 (-)} & \multicolumn{1}{l|}{\textbf{0.707 (-)}} & \textbf{0.750 (-)} & \multicolumn{1}{l|}{\textbf{0.707 (-)}} & \textbf{0.750} & \multicolumn{1}{l|}{\textbf{0.707}} & 0.542 & 0.118 \\
D15 & 0.500 (=) & \multicolumn{1}{l|}{0.000 (+)} & 0.492 (+) & \multicolumn{1}{l|}{0.000 (+)} & 0.411 (+) & \multicolumn{1}{l|}{0.000 (+)} & 0.492 (+) & \multicolumn{1}{l|}{0.000 (+)} & 0.500 (=) & \multicolumn{1}{l|}{0.000 (+)} & 0.492 (+) & \multicolumn{1}{l|}{0.000 (+)} & 0.492 (+) & \multicolumn{1}{l|}{0.000 (+)} & \textbf{0.521} & \multicolumn{1}{l|}{\textbf{0.417}} & 0.499 & 0.032 \\
D16 & 0.500 (=) & \multicolumn{1}{l|}{0.000 (=)} & 0.504 (-) & \multicolumn{1}{l|}{0.332 (-)} & \textbf{0.521 (-)} & \multicolumn{1}{l|}{\textbf{0.339 (-)}} & 0.460 (+) & \multicolumn{1}{l|}{0.000 (=)} & 0.453 (+) & \multicolumn{1}{l|}{0.000 (=)} & 0.498 (+) & \multicolumn{1}{l|}{0.330 (-)} & 0.505 (-) & \multicolumn{1}{l|}{0.333 (-)} & 0.503 & \multicolumn{1}{l|}{0.110} & 0.498 & 0.009 \\
D17 & 0.750 (+) & \multicolumn{1}{l|}{0.707 (=)} & \textbf{0.875 (-)} & \multicolumn{1}{l|}{\textbf{0.866 (-)}} & 0.938 (-) & \multicolumn{1}{l|}{0.935 (-)} & 0.500 (+) & \multicolumn{1}{l|}{0.000 (+)} & 0.500 (+) & \multicolumn{1}{l|}{0.000 (+)} & \textbf{0.875 (-)} & \multicolumn{1}{l|}{\textbf{0.866 (-)}} & \textbf{0.875 (-)} & \multicolumn{1}{l|}{\textbf{0.866 (-)}} & \textbf{0.875} & \multicolumn{1}{l|}{\textbf{0.866}} & 0.781 & 0.731 \\
D18 & 0.500 (+) & \multicolumn{1}{l|}{0.000 (+)} & 0.488 (+) & \multicolumn{1}{l|}{0.000 (+)} & 0.488 (+) & \multicolumn{1}{l|}{0.000 (+)} & 0.498 (+) & \multicolumn{1}{l|}{0.000 (+)} & 0.498 (+) & \multicolumn{1}{l|}{0.000 (+)} & 0.476 (+) & \multicolumn{1}{l|}{0.000 (+)} & 0.488 (+) & \multicolumn{1}{l|}{0.000 (+)} & \textbf{0.585} & \multicolumn{1}{l|}{\textbf{0.413}} & 0.506 & 0.044 \\
D19 & 0.559 (-) & \multicolumn{1}{l|}{0.352 (-)} & 0.555 (+) & \multicolumn{1}{l|}{0.351 (+)} & 0.554 (+) & \multicolumn{1}{l|}{0.351 (+)} & 0.496 (+) & \multicolumn{1}{l|}{0.000 (+)} & 0.499 (+) & \multicolumn{1}{l|}{0.000 (+)} & 0.551 (+) & \multicolumn{1}{l|}{0.350 (+)} & 0.555 (+) & \multicolumn{1}{l|}{0.351 (+)} & \textbf{0.670} & \multicolumn{1}{l|}{\textbf{0.602}} & 0.568 & 0.369 \\
D20 & 0.500 (+) & \multicolumn{1}{l|}{0.000 (+)} & 0.898 (+) & \multicolumn{1}{l|}{0.892 (+)} & 0.600 (+) & \multicolumn{1}{l|}{0.447 (+)} & 0.500 (+) & \multicolumn{1}{l|}{0.000 (+)} & 0.500 (+) & \multicolumn{1}{l|}{0.000 (+)} & 0.897 (+) & \multicolumn{1}{l|}{0.891 (+)} & 0.898 (+) & \multicolumn{1}{l|}{0.892 (+)} & \textbf{1.000} & \multicolumn{1}{l|}{\textbf{1.000}} & 0.890 & 0.866 \\ \hline
Overall & \multicolumn{18}{c|}{172 +, 78 -, 30 =} \\ \hline
\end{tabular}
}
\end{minipage}
\end{table*}

TABLES~\ref{tab:ADA_result},~\ref{tab:GBDT_result},~\ref{tab:RF_result} and~\ref{tab:SVM_result} report the AUC and G\_Mean results of the 4 classification algorithms in the test sets after the training sets rebalanced by EvoSampling and the baseline sampling methods. 
In the four tables, ``+'', ``-'', and ``='' indicate that EvoSampling is significantly better, worse, or no significant difference from a baseline method, respectively. 
The ``best'' column under EvoSampling represents the highest AUC/G\_Mean result obtained from the 30 runs on each dataset, while the ``mean'' column shows the average AUC/G\_Mean result across the 30 runs.

The column named ``Initial'' in TABLES~\ref{tab:ADA_result},~\ref{tab:GBDT_result},~\ref{tab:RF_result} and~\ref{tab:SVM_result} indicates the initial results of the four classification algorithms without using any sampling methods to rebalance the training sets. According to the results of statistical significance tests, after using EvoSampling, AdaBoost/GBDT/RF/SVMs achieve significantly better AUC/G\_Mean performances than those using the original data in 55/58 out of the total 80 cases across the 20 datasets. These results demonstrate the effectiveness of our method in generating well-balanced datasets with high-quality instances. Moreover, compared with the baseline methods, EvoSampling assists AdaBoost/GBDT/RF/SVMs to achieve better AUC and G\_Mean performance on the 20 datasets. It can be found from TABLES~\ref{tab:ADA_result},~\ref{tab:GBDT_result},~\ref{tab:RF_result} that EvoSampling can assist the ensemble learning-based classifiers (i.e., AdaBoost, GBDT, and RF) to achieve significantly better AUC/G\_Mean performance than using other baseline sampling methods in 261/282 out of the 420 cases. According to TABLE~\ref{tab:SVM_result}, EvoSampling can assist SVMs to achieve significantly better AUC/G\_Mean performance than other baseline sampling methods in 82/90 out of the 140 cases. Notably, on the highly imbalanced datasets with $IR > 60$ (such as winequality-red-3\_vs\_5, poker-8-9\_vs\_5, and poker-8\_vs\_6), EvoSampling shows a significant superiority over the initial data and the baseline sampling methods in 55/59 (AUC/G\_Mean) out of a total 84 cases.


The goal of sampling methods is to create a single balanced dataset by generating a set of high-quality instances for the minority class. With this in mind, we compare the best results of EvoSampling across the 30 runs with those of a baseline.
Based on the best results from the 30 runs of EvoSampling, EvoSampling enables AdaBoost/GBDT/RF/SVMs to achieve the highest AUC performance on 19/20/17/18 out of the 20 datasets,  and the highest G\_Mean performance on 18/19/17/18 out of the 20 datasets, compared to the baseline methods.

\subsection{Summary on the Results and Further Analysis}

TABLE \ref{tab:main_results} summarizes the overall test results (AUC and G\_Mean) of the 4 classifiers, trained individually on the original data, rebalanced data by EvoSampling and the six baseline methods across the 20 datasets. In TABLE~\ref{tab:main_results}, ``Avg.AUC'' and ``Avg.G\_Mean'' respectively represent the average AUC and G\_Mean performances of the 4 classifiers on the original data (named ``initial'') or with each sampling method across the 20 datasets. 
``Avg.RANK\_1'' represents the average rank of EvoSampling's best results based on a corresponding classification metric (AUC or G\_Mean) compared to the baseline methods. The Avg.RANK\_1 of each method (e.g., SMOTE or ADASYN) is calculated as follows. First, the 8 methods (Initial, the 6 baseline methods, and EvoSampling) are ranked according to their AUC or G\_Mean performance of a classifier on a dataset. Each method receives 80 (20 datasets $\times$ 4 classifiers) rankings, with ranking values ranging from 1 to 8. These rankings are summed for each method and then averaged to determine the Avg.RANK\_1 of this method. ``Avg.RANK\_2'' represents the average rank of EvoSampling's mean results (rather than the best result) based on a performance metric compared to the baseline methods.

As shown in TABLE \ref{tab:main_results}, for both the AUC and G\_Mean performance, the best result of EvoSampling is better than the other methods in terms of Avg.AUC and Avg. G\_Mean, and the mean result of EvoSampling ranks second. 
More importantly, EvoSampling ranks the first in terms of Avg.RANK\_1 and Avg.RANK\_2. 
These results indicate that EvoSampling is a competitive sampling method that effectively rebalances data by generating a set of good-quality instances and removing low-quality ones. 

\begin{table*}[]
\renewcommand\arraystretch{1.2}
\caption{
Summaries of the results on the 20 test sets. The best results are in \textbf{bold}, and the second best are {\ul underlined}.
}
\label{tab:main_results}
\resizebox{\linewidth}{!}
{
\begin{tabular}{|cccccccccc|}
\hline
\multicolumn{1}{|c|}{\multirow{2}{*}{Metric}} & \multirow{2}{*}{Initial} & \multirow{2}{*}{SMOTE} & \multirow{2}{*}{ADASYN} & \multirow{2}{*}{Borderline-SMOTE1} & \multirow{2}{*}{Borderline-SMOTE2} & \multirow{2}{*}{SMOTE+ENN} & \multicolumn{1}{c|}{\multirow{2}{*}{SMOTE+Tomek}} & \multicolumn{2}{c|}{\textbf{EvoSampling}} \\ \cline{9-10} 
\multicolumn{1}{|c|}{} &  &  &  &  &  &  & \multicolumn{1}{c|}{} & best & mean \\ \hline
\multicolumn{10}{|c|}{AUC} \\ \hline
\multicolumn{1}{|c|}{Avg.AUC} & 0.641 & 0.680 & 0.670 & 0.653 & 0.648 & 0.699 & \multicolumn{1}{c|}{0.677} & \textbf{0.830} & {\ul 0.717} \\
\multicolumn{1}{|c|}{Avg.Rank\_1} & 4.950 & {\ul  4.694} & 5.075 & 5.219 & 5.131 & 4.706 & \multicolumn{1}{c|}{4.844} & \textbf{1.381} & - \\
\multicolumn{1}{|c|}{Avg.Rank\_2} & 4.763 & 4.444 & 4.813 & 5.025 & 4.900 & {\ul 4.313} & \multicolumn{1}{c|}{4.619} & - & \textbf{3.125} \\ \hline
\multicolumn{10}{|c|}{G\_Mean} \\ \hline
\multicolumn{1}{|c|}{Avg.G\_Mean} & 0.383 & 0.495 & 0.499 & 0.433 & 0.414 & 0.587 & \multicolumn{1}{c|}{0.506} & \textbf{0.802} & {\ul 0.618} \\
\multicolumn{1}{|c|}{Avg.Rank\_1} & 5.406 & 4.619 & 4.831 & 5.294 & 5.525 & {\ul 4.225} & \multicolumn{1}{c|}{4.681} & \textbf{1.419} & - \\
\multicolumn{1}{|c|}{Avg.Rank\_2} & 5.25 & 4.363 & 4.531 & 5.088 & 5.344 & {\ul 3.856} & \multicolumn{1}{c|}{4.431} & - & \textbf{3.138} \\ \hline
\end{tabular}
}
\end{table*}

\begin{table*}[]
\caption{Summaries of the results about a specific classification algorithm on the 20 test sets. The best results are in \textbf{bold}, and the second best are {\ul underlined}.}
\centering
\renewcommand\arraystretch{1.2}
\label{tab:auc_results}
\noindent
\begin{minipage}{1\textwidth}  
  \raggedright  
  \resizebox{\textwidth}{!}{%
\begin{tabular}{ccccccccc}
\hline
\multicolumn{1}{|c|}{\multirow{2}{*}{Method}} & \multicolumn{2}{c|}{AdaBoost} & \multicolumn{2}{c|}{GBDT} & \multicolumn{2}{c|}{RF} & \multicolumn{2}{c|}{SVMs} \\ \cline{2-9} 
\multicolumn{1}{|c|}{} & \multicolumn{1}{c|}{Avg.Rank} & \multicolumn{1}{c|}{Number (+/=/-)} & \multicolumn{1}{c|}{Avg.Rank} & \multicolumn{1}{c|}{Number (+/=/-)} & \multicolumn{1}{c|}{Avg.Rank} & \multicolumn{1}{c|}{Number (+/=/-)} & \multicolumn{1}{c|}{Avg.Rank} & \multicolumn{1}{c|}{Number (+/=/-)} \\ \hline
\multicolumn{9}{|c|}{AUC} \\ \hline
\multicolumn{1}{|c|}{Initial} & \multicolumn{1}{c|}{{\ul 4.075}} & \multicolumn{1}{c|}{13/3/4} & \multicolumn{1}{c|}{5.275} & \multicolumn{1}{c|}{14/1/5} & \multicolumn{1}{c|}{5.600} & \multicolumn{1}{c|}{17/0/3} & \multicolumn{1}{c|}{4.100} & \multicolumn{1}{c|}{11/3/6} \\
\multicolumn{1}{|c|}{SMOTE} & \multicolumn{1}{c|}{5.300} & \multicolumn{1}{c|}{15/4/1} & \multicolumn{1}{c|}{{\ul 4.150}} & \multicolumn{1}{c|}{8/4/8} & \multicolumn{1}{c|}{{\ul 4.250}} & \multicolumn{1}{c|}{11/3/6} & \multicolumn{1}{c|}{{\ul 4.075}} & \multicolumn{1}{c|}{10/2/8} \\
\multicolumn{1}{|c|}{ADASYN} & \multicolumn{1}{c|}{4.700} & \multicolumn{1}{c|}{14/3/3} & \multicolumn{1}{c|}{4.625} & \multicolumn{1}{c|}{11/2/7} & \multicolumn{1}{c|}{4.575} & \multicolumn{1}{c|}{12/2/6} & \multicolumn{1}{c|}{5.350} & \multicolumn{1}{c|}{13/1/6} \\
\multicolumn{1}{|c|}{Borderline-SMOTE1} & \multicolumn{1}{c|}{5.200} & \multicolumn{1}{c|}{16/1/3} & \multicolumn{1}{c|}{4.575} & \multicolumn{1}{c|}{10/3/7} & \multicolumn{1}{c|}{4.650} & \multicolumn{1}{c|}{14/0/6} & \multicolumn{1}{c|}{5.675} & \multicolumn{1}{c|}{14/2/4} \\
\multicolumn{1}{|c|}{Borderline-SMOTE2} & \multicolumn{1}{c|}{4.475} & \multicolumn{1}{c|}{15/3/2} & \multicolumn{1}{c|}{4.800} & \multicolumn{1}{c|}{12/2/6} & \multicolumn{1}{c|}{5.100} & \multicolumn{1}{c|}{15/1/4} & \multicolumn{1}{c|}{5.225} & \multicolumn{1}{c|}{13/3/4} \\
\multicolumn{1}{|c|}{SMOTE+ENN} & \multicolumn{1}{c|}{4.450} & \multicolumn{1}{c|}{11/3/6} & \multicolumn{1}{c|}{4.375} & \multicolumn{1}{c|}{7/2/11} & \multicolumn{1}{c|}{ 4.275} & \multicolumn{1}{c|}{11/2/7} & \multicolumn{1}{c|}{4.15} & \multicolumn{1}{c|}{10/4/6} \\
\multicolumn{1}{|c|}{SMOTE+Tomek} & \multicolumn{1}{c|}{5.550} & \multicolumn{1}{c|}{16/3/1} & \multicolumn{1}{c|}{4.400} & \multicolumn{1}{c|}{8/4/8} & \multicolumn{1}{c|}{4.300} & \multicolumn{1}{c|}{11/3/6} & \multicolumn{1}{c|}{4.225} & \multicolumn{1}{c|}{11/2/7} \\
\multicolumn{1}{|c|}{\textbf{EvoSampling}} & \multicolumn{1}{c|}{\textbf{2.250}} & \multicolumn{1}{c|}{-/-/-} & \multicolumn{1}{c|}{\textbf{3.800}} & \multicolumn{1}{c|}{-/-/-} & \multicolumn{1}{c|}{\textbf{3.250}} & \multicolumn{1}{c|}{-/-/-} & \multicolumn{1}{c|}{\textbf{3.200}} & \multicolumn{1}{c|}{-/-/-} \\ \hline
\multicolumn{9}{|c|}{G\_Mean} \\ \hline
\multicolumn{1}{|c|}{Initial} & \multicolumn{1}{c|}{4.950} & \multicolumn{1}{c|}{13/4/3} & \multicolumn{1}{c|}{5.425} & \multicolumn{1}{c|}{15/0/5} & \multicolumn{1}{c|}{5.875} & \multicolumn{1}{c|}{18/0/2} & \multicolumn{1}{c|}{4.750} & \multicolumn{1}{c|}{12/3/5} \\
\multicolumn{1}{|c|}{SMOTE} & \multicolumn{1}{c|}{4.950} & \multicolumn{1}{c|}{16/2/2} & \multicolumn{1}{c|}{4.125} & \multicolumn{1}{c|}{11/2/7} & \multicolumn{1}{c|}{4.300} & \multicolumn{1}{c|}{13/3/4} & \multicolumn{1}{c|}{4.075} & \multicolumn{1}{c|}{12/1/7} \\
\multicolumn{1}{|c|}{ADASYN} & \multicolumn{1}{c|}{4.225} & \multicolumn{1}{c|}{13/5/2} & \multicolumn{1}{c|}{4.525} & \multicolumn{1}{c|}{12/1/7} & \multicolumn{1}{c|}{4.475} & \multicolumn{1}{c|}{13/3/4} & \multicolumn{1}{c|}{4.900} & \multicolumn{1}{c|}{13/2/5} \\
\multicolumn{1}{|c|}{Borderline-SMOTE1} & \multicolumn{1}{c|}{5.550} & \multicolumn{1}{c|}{16/2/2} & \multicolumn{1}{c|}{4.700} & \multicolumn{1}{c|}{12/2/6} & \multicolumn{1}{c|}{4.700} & \multicolumn{1}{c|}{14/1/5} & \multicolumn{1}{c|}{5.400} & \multicolumn{1}{c|}{14/3/3} \\
\multicolumn{1}{|c|}{Borderline-SMOTE2} & \multicolumn{1}{c|}{5.125} & \multicolumn{1}{c|}{14/5/1} & \multicolumn{1}{c|}{5.250} & \multicolumn{1}{c|}{14/2/4} & \multicolumn{1}{c|}{5.550} & \multicolumn{1}{c|}{17/0/3} & \multicolumn{1}{c|}{5.450} & \multicolumn{1}{c|}{15/1/4} \\
\multicolumn{1}{|c|}{SMOTE+ENN} & \multicolumn{1}{c|}{{\ul 3.600}} & \multicolumn{1}{c|}{11/1/8} & \multicolumn{1}{c|}{{\ul 4.075}} & \multicolumn{1}{c|}{8/2/10} & \multicolumn{1}{c|}{{\ul 3.900}} & \multicolumn{1}{c|}{12/4/4} & \multicolumn{1}{c|}{{\ul 3.850}} & \multicolumn{1}{c|}{11/2/7} \\
\multicolumn{1}{|c|}{SMOTE+Tomek} & \multicolumn{1}{c|}{4.950} & \multicolumn{1}{c|}{17/2/1} & \multicolumn{1}{c|}{4.300} & \multicolumn{1}{c|}{10/3/7} & \multicolumn{1}{c|}{4.250} & \multicolumn{1}{c|}{13/3/4} & \multicolumn{1}{c|}{4.225} & \multicolumn{1}{c|}{13/1/6} \\
\multicolumn{1}{|c|}{\textbf{EvoSampling}} & \multicolumn{1}{c|}{\textbf{2.650}} & \multicolumn{1}{c|}{-/-/-} & \multicolumn{1}{c|}{\textbf{3.600}} & \multicolumn{1}{c|}{-/-/-} & \multicolumn{1}{c|}{\textbf{2.950}} & \multicolumn{1}{c|}{-/-/-} & \multicolumn{1}{c|}{\textbf{3.335}} & \multicolumn{1}{c|}{-/-/-} \\ \hline
 &  &  &  &  &  &  &  & 
\end{tabular}
}
\end{minipage}
\end{table*}

To investigate whether EvoSampling can develop good-quality balanced datasets to improve the performance of a classifier, we summarize the average rankings (Avg.Rank) of the different methods on each classification algorithm across the 20 datasets in TABLE ~\ref{tab:auc_results}. 
As shown in TABLE ~\ref{tab:auc_results}, based on the AUC results, EvoSampling achieves Avg.Rank of 2.250, 3.800, 3.250, and 3.200 on AdaBoost, GBDT, RF, and SVMs, respectively. It outperforms the other seven methods, particularly excelling with AdaBoost, where it significantly surpasses the second-best method. Similarly, for the G\_Mean results, EvoSampling achieves Avg.Rank of 2.650, 3.600, 2.950, and 3.335 on AdaBoost, GBDT, RF, and SVMs, outperforming all other methods. These results indicate our proposed EvoSampling method can benefit different classifiers to improve their performance for imbalanced classification. 

According to the summarized significance results in TABLE ~\ref{tab:auc_results}, compared with original data or the baseline sampling methods, EvoSampling can help AdaBoost/GBDT/RF/SVMs achieve significantly better or similar AUC performance in 120/88/102/99 out of the 140 cases. Similarly, EvoSampling can help AdaBoost/GBDT/RF/SVMs achieve significantly better or similar G\_Mean performance in 121/94/114/103 out of the 140 cases. These results demonstrate that EvoSampling surpasses the baseline sampling methods in enhancing data quality for classifiers, particularly for those based on ensemble learning.
Ensemble classifiers typically perform well with diverse data. Thus, one key reason for EvoSampling's strong performance with ensemble classifiers is its ability to generate more diverse instances than the baseline methods across the datasets in the experiments.

\subsection{Ablation Studies}
\begin{figure*}
	\centering
	\includegraphics[ width=1\textwidth]{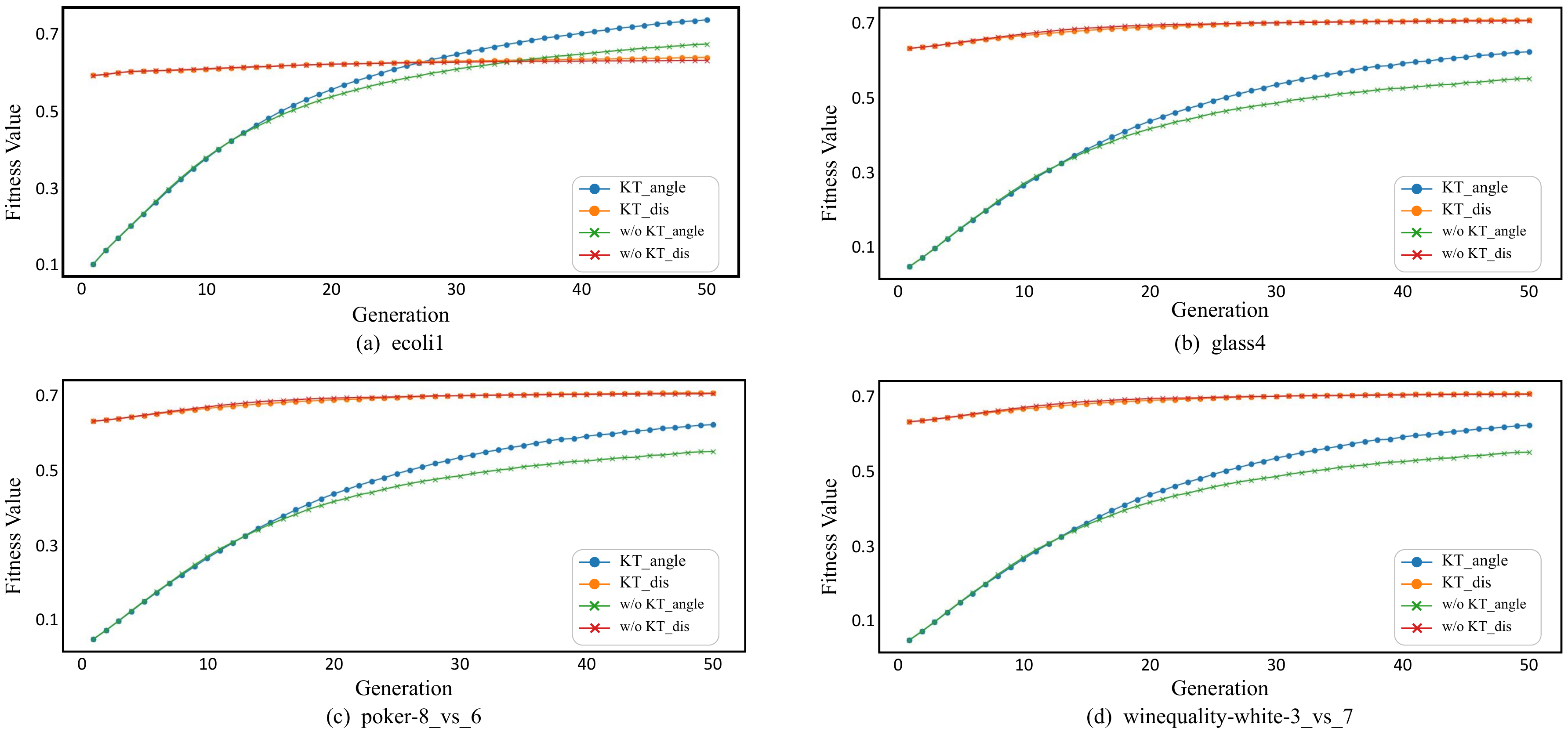}
	\caption{The results of the ablation study on the 4 datasets.} 
	\label{fig:ablation}
\end{figure*}

We further investigate the impact of knowledge transfer (KT)
during the GP evolutionary process in EvoSampling. We conducted an ablation study using 4 datasets, with significantly different $IR$s (where the dataset ecoli-0-6-7\_vs\_3-5 with the $IR$ of 9.09; the dataset glass4 with the $IR$ of 15.47; the dataset winequality-white-3\_vs\_7 with the $IR$ of 44; the dataset poker-8\_vs\_6 with the $IR$ of 85.88). In the ablation study, we remove the knowledge transfer module, allowing $n$ GP process to evolve independently through standard crossover and mutation operations. We refer to this ablation method as w/o KT, and plot the convergence curves of the two measures (i.e., the distance $D$ and angle $\theta$) in the proposed fitness function during the evolutionary phase for both EvoSampling with KT and w/o KT in Fig.~\ref{fig:ablation}. 
It is evident that the angle values for w/o KT are consistently lower in the same generation. These findings indicate that knowledge transfer effectively accelerates the GP evolutionary process.


\section{Conclusions}
This study aims to generate diverse and high-quality instances while effectively removing noise to enhance the data quality in imbalanced classification. The goal has been successfully achieved by proposing an evolutionary hybrid multi-granularity sampling method. 
Specifically, we have designed a multi-task GP-based oversampling method that incorporates a knowledge transfer mechanism to enhance performance. 
Furthermore, GBC has been applied to perform undersampling on the oversampled data to remove low-quality instances.

Experimental results on the 20 imbalanced datasets have demonstrated that the proposed EvoSampling method successfully generated higher-quality datasets, helping classification algorithms achieve superior performance over the baseline methods. 
In addition, the ablation studies have indicated that the knowledge transfer mechanism effectively accelerated the evolutionary speed of GP. Although EvoSampling is effective in generating diverse and high-quality instances, the multi-task GP requires significant computational resources and consumes long training time. In the future, we will explore how to accelerate the process.

\ifCLASSOPTIONcaptionsoff
  \newpage
\fi

\bibliographystyle{IEEEtran}
\bibliography{reference}




\begin{IEEEbiography}[{\includegraphics[width=0.9in,height=1.25in]{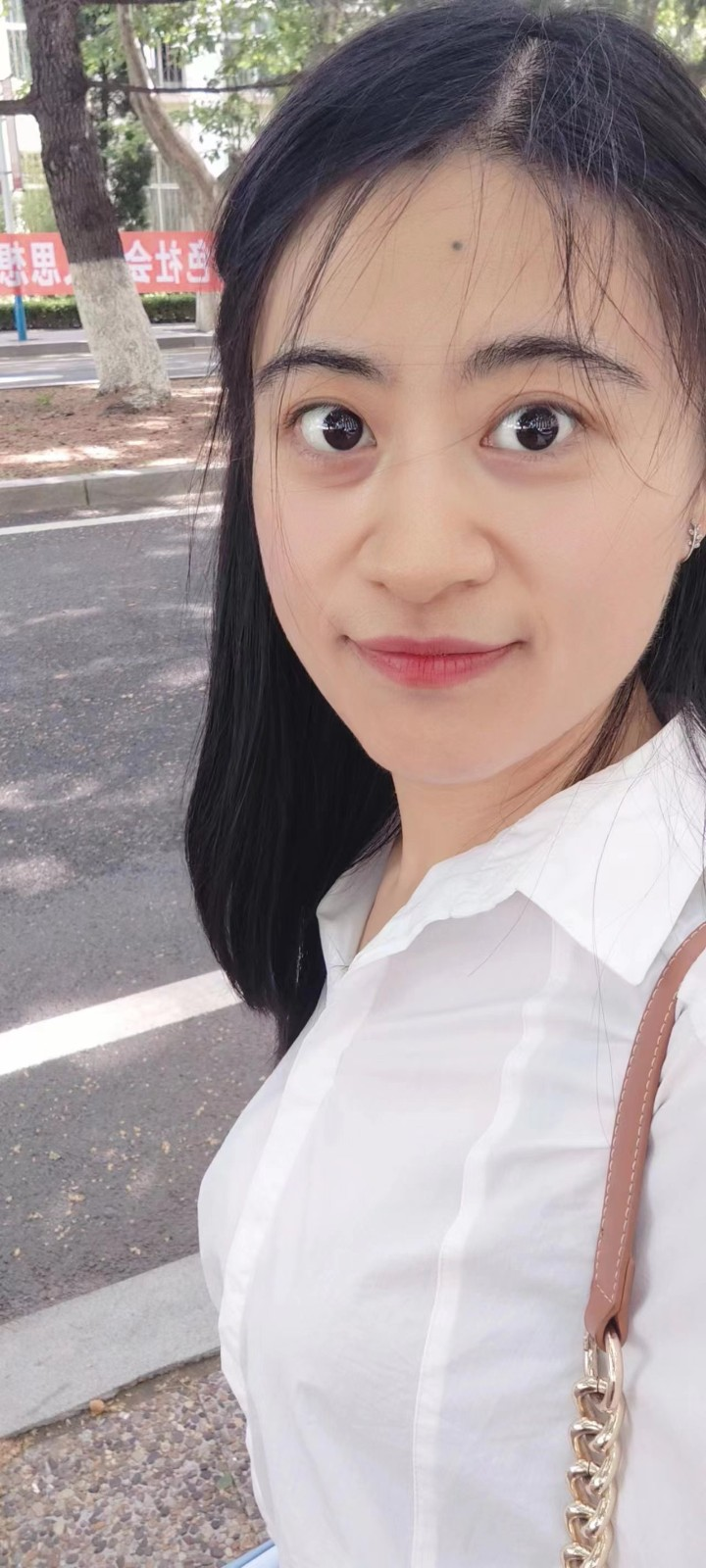}}]{Wenbin Pei} received her Ph.D. degree at Victoria University of Wellington, New Zealand, in 2021. She is currently an assistant professor at Dalian University of Technology. Her research interests include evolutionary computation and machine learning. She has published more than 30 papers in fully refereed international journals and conferences, including IEEE Transactions on Evolutionary Computation, Evolutionary Computation Journal (MIT Press), IEEE Computational Intelligence Magazine, and AAAI. She was a program committee member for 6 international conferences and a co-chair of special sessions in 4 international conferences. She has been serving as a reviewer for international journals, including IEEE Transactions on Evolutionary Computation, IEEE Transactions on Cybernetics, Evolutionary Computation Journal (MIT Press), IEEE Transactions on Emerging Topics in Computational Intelligence, Applied Soft Computing, and Artificial Intelligence in Medicine. %
\end{IEEEbiography}

\begin{IEEEbiography}[{\includegraphics[width=1in,height=1.4in]{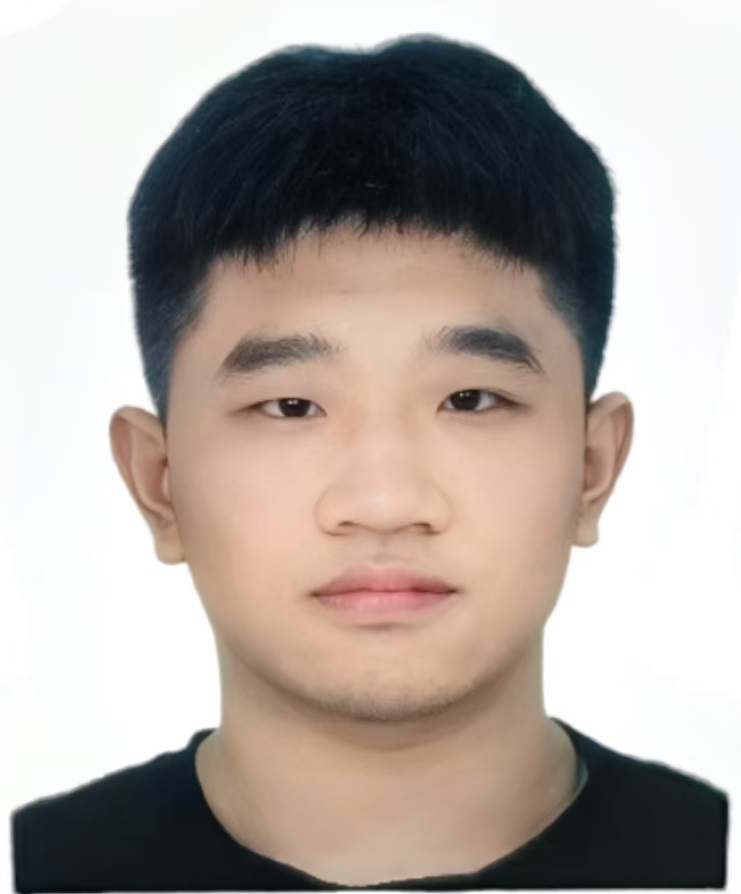}}]{Ruohao Dai} received his Bachelor at Hebei Unversity, China, in 2023. He is currently pursuing his Master's degree at the School of Computer Science and Technology, Dalian University of Technology. His research interests include evolutionary computation, imbalanced learning, and time-series analysis. %
\end{IEEEbiography}

\begin{IEEEbiography}[{\includegraphics[width=1.05in,height=1.2in]{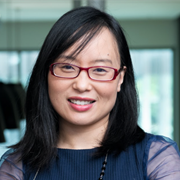}}]{Bing Xue} 
(Fellow, IEEE) received her Ph.D. degree at Victoria University of Wellington, New Zealand, in 2014. She is a fellow of Engineering New Zealand. She is currently a Professor and the deputy head of the School of Engineering and Computer Science at Victoria University of Wellington. Her research focuses mainly on evolutionary computation, feature learning, and machine learning. She has more than 300 papers published in fully refereed international journals and conferences. Prof. Xue has also served as Associate Editor or Member of the Editorial Board for 10 international journals, including IEEE Transactions on Evolutionary Computation, IEEE Transactions on Artificial Intelligence, and IEEE Computational Intelligence Magazine. 
\end{IEEEbiography}

\vspace{-11mm} 
\begin{IEEEbiography}[{\includegraphics[width=1in,height=1.25in]{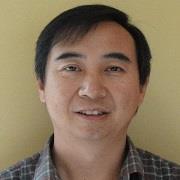}}]{Mengjie Zhang} (Fellow, IEEE) received 
his Ph.D degree in Computer Science from RMIT University, Australia in 2000. He is a fellow of the Royal Society of New Zealand and a fellow of Engineering New Zealand. He is currently a Professor of Computer Science, the Head of the Evolutionary Computation and Machine Learning Research Group, and the Associate Dean (Research and Innovation) with the Faculty of Engineering, at Victoria University of Wellington, New Zealand. His research interests include genetic programming, image analysis, feature selection and reduction, job-shop scheduling, evolutionary deep learning, and transfer learning. He has published over 800 research papers in refereed international journals and conferences. He has been serving as an associated editor or editorial board member for over 10 international journals, including IEEE Transactions on Evolutionary Computation, IEEE Transactions on Cybernetics, IEEE Transactions on Emergent Topics in Computational Intelligence, and Evolutionary Computation (MIT Press). 
\end{IEEEbiography}

\begin{IEEEbiography}[{\vspace{-4mm} \includegraphics[width=1in,height=1.27in]{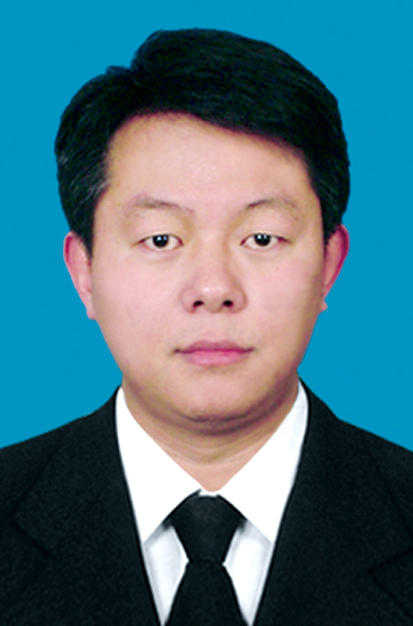}}]{Qiang Zhang} received the Ph.D. degree in Circuits and Systems from Xidian University, Xi'an, in 2002. He is currently a professor and the dean of the School of Computer Science and Technology, at Dalian University of Technology. His research interests include bio-inspired computing and related applications. Prof. Zhang has published more than 70 papers in fully refereed international journals and conferences. He was awarded the National Science Fund for Distinguished Young Scholars in 2014, and was also selected as one of the State Department special allowance experts. He has been serving as editorial board for 7 international journals and chairs of special issues in journals, such as Neurocomputing and the International Journal of Computer Applications in Technology.
\end{IEEEbiography}

\begin{IEEEbiography}[{\vspace{-4mm} \includegraphics[width=1in,height=1.2in]{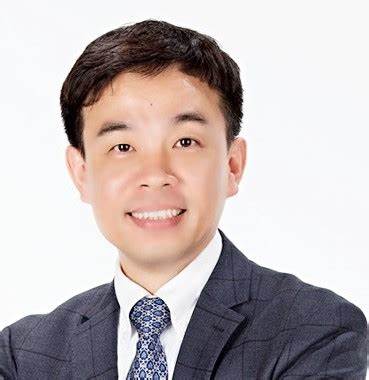}}]{Yiu-Ming Cheung} (Fellow, IEEE) received the Ph.D. degree from the Department of Computer Science and Engineering, the Chinese University of Hong Kong, Hong Kong, in 2000. He is currently a Chair Professor with the Department of Computer Science, Hong Kong Baptist University, Hong Kong. His research interests include machine learning and visual computing, as well as, data science, pattern recognition and multi-objective optimization. Dr. Cheung is also a fellow of AAAS, IAPR, IET, and BCS. He is currently the Editor-in-chief of IEEE Transactions on Emerging Topics in Computational Intelligence. Also, he has served as an Associate Editor for several prestigious journals, including IEEE Transactions on Cybernetics, IEEE Transactions on Cognitive and Developmental Systems, IEEE Transactions on Neural Networks and Learning Systems, Pattern Recognition, and so on.
\end{IEEEbiography}

\begin{IEEEbiography}[{\vspace{-4mm} \includegraphics[width=1.3in,height=1.1in]{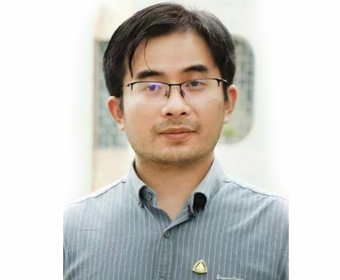}}]{Shuyin Xia} received the BS and MS degrees in computer science from the Chongqing University of Technology, China, in 2008 and 2012, and the PhD degree from the College of Computer Science, Chongqing University, in China. He is a professor and a PhD supervisor with the CQUPT (Chongqing University
of Posts and Telecommunications) in Chongqing,
China. He is also a distinguished professor of national scholars in CQUPT. His research interests include granular computing, data mining, and machine
learning.
\end{IEEEbiography}

\end{document}